\documentclass[runningheads,dvipsnames,table]{llncs}

\usepackage[mobile]{eccv}

\usepackage{eccvabbrv}

\usepackage{graphicx}
\usepackage{booktabs}
\usepackage{multirow}

\usepackage{enumitem}
\usepackage{pifont}% http://ctan.org/pkg/pifont  % for cmark, xmark
\usepackage{afterpage}
\usepackage{etoolbox}  % for teaser

\definecolor{tabfirst}{rgb}{0.5, 1.0, 0.5} 
\definecolor{tabsecond}{rgb}{1, 1, 0.5}  

\newcommand{\cmark}{\ding{51}}%
\newcommand{\xmark}{\ding{55}}%

\newcommand{\Indexed}[1]{#1^{n}}
\newcommand{\Setof}[1]{\{#1^{n}\}}
\newcommand{\HAMER}{HaMeR}
\usepackage[accsupp]{axessibility}  % Improves PDF readability for those with disabilities.

\usepackage[pagebackref,breaklinks,colorlinks]{hyperref}
\usepackage{orcidlink}

\begin{document}

% ---------------------------------------------------------------
\title{Get a Grip: Reconstructing Hand-Object Stable Grasps in Egocentric Videos} 

\author{Zhifan Zhu \qquad \qquad Dima Damen\\
}
\institute{
School of Computer Science, University of Bristol, UK\\
\vspace*{6pt}
\url{https://zhifanzhu.github.io/getagrip}
}

\authorrunning{Z. Zhu and D. Damen}

\maketitle

\begin{abstract}
We propose the task of Hand-Object Stable Grasp Reconstruction (HO-SGR), the reconstruction of frames during which the hand is stably holding the object. 
We first develop the stable grasp definition based on the intuition that the
in-contact area between the hand and object should remain stable.
By analysing the 3D ARCTIC dataset, 
we identify stable grasp durations and showcase that objects in stable grasps move within a single degree of freedom (1-DoF).
We thereby propose a method to jointly optimise all frames within a stable grasp, minimising object motions to a latent 1-DoF.
Finally, we extend the knowledge to in-the-wild videos by labelling 2.4K clips of stable grasps.
Our proposed EPIC-Grasps dataset includes 390 object instances of 9 categories, featuring stable grasps from videos of daily interactions in 141 environments.
Without 3D ground truth, we use stable contact areas and 2D projection masks to assess the HO-SGR task in the wild.
We evaluate relevant methods and our approach preserves significantly higher stable contact area,
on both EPIC-Grasps and stable grasp sub-sequences from the ARCTIC dataset.
%See project page at .
\end{abstract}

\section{Introduction}
\label{sec:intro}

Accurately reconstructing three-dimensional hands along with a grasped or manipulated object is key to unlocking many perception problems, including fine-grained understanding of interactions, but also potential applications including augmented reality, robotic imitation learning and human-machine interactions.

Prior approaches have laid the foundation for the task, though focusing mostly on 3D input~\cite{TaheriGRAB:Objects, BrahmbhattContactPose:Pose, Zhang2021ManipNet:Representation}, which limits its application to the vast amount of video footage. Approaches with 2D input have relied on 3D ground truth for supervision~\cite{YeWhatsHands, Chen2022AlignSDF:Reconstruction, Hasson2021TowardsVideos, Karunratanakul2020GraspingGrasps, Yang2020CPF:Interaction}, again restricting their scope to curated setups where multi-camera sensing and 3D ground truth can be obtained.
Early insights~\cite{Hasson2021TowardsVideos, CaoReconstructingWild, Patel2022LearningVideos} demonstrate that evaluation on in-the-wild egocentric footage remains significantly challenging.

In this work, we focus on the task of reconstruction in-the-wild on temporal periods of stable grasps. 
Our contribution is composed of three components.
\textbf{First,} we propose the task of Hand-Object Stable Grasp Reconstruction (HO-SGR) which jointly optimises the reconstructions across all frames within one stable grasp. We showcase that objects move within one degree of freedom (1-DoF), relative to the hand pose, throughout the stable grasp.
\textbf{Second,} we accordingly propose a method that jointly reconstructs the hands and objects by minimising the object's motion, relative to the hand, to 1-DoF around a latent rotation axis, throughout the frames.
We demonstrate our method outperforms baselines and alternative assumptions of object movement using 3D ground truth from the stable grasps within the egocentric views of the ARCTIC dataset~\cite{fan2023arctic}.
\textbf{Third}, We label a sizeable dataset of 2.4K stable grasps clips from egocentric videos.
Our EPIC-Grasps dataset is the first for hand-object reconstruction collected from unscripted activities, with individuals grasping 390 different objects by both hands. 
Similar to previous works \cite{HampaliHOnnotate:Poses, ChaoDexYCB:Objects, Garcia-HernandoFirst-PersonAnnotations, CaoReconstructingWild, Patel2022LearningVideos}, we restrict our evaluation to known category CAD models.
Our dataset comes with pseudo-ground truth in the form of 2D segmentation masks available from~\cite{VISOR2022}, allowing to measure the 3D reconstruction's projection relative to this 2D ground truth.

We evaluate our proposed method on EPIC-Grasps, outperforming other baselines using the proxy measure of stable grasps within 2D projections.
We demonstrate that joint optimisation within the stable grasp is vital in the egocentric setup as finger joints are often concealed from the camera viewpoint introducing additional ambiguity when reconstructing from a single image. 

\begin{figure}[t]
    \centering
    \includegraphics[width=\linewidth]{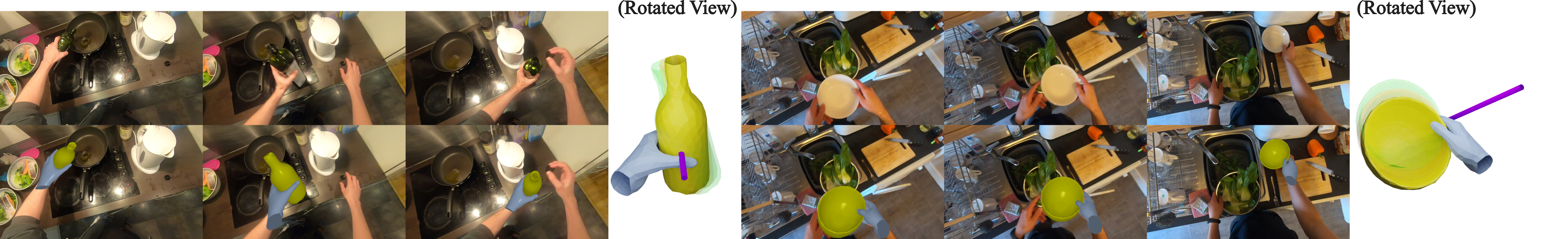}
    \caption{
    Two stable grasp sequences from EPIC-Grasps for a bottle (left) and bowl (right). We show sample frames (top) and reconstructions (bottom). Right: for each reconstruction, we show the rotated view, along with the latent 1-DoF axis.
    }
    \label{fig:teaser}
\end{figure}

We demonstrate a sample from our EPIC-Grasps dataset along with reconstructions from our method in~\cref{fig:teaser}.

\section{Related Works}
\label{sec:related}

For comparison of hand object reconstruction datasets, see~\cref{sec:dataset}.
We review related works in hand pose estimation and hand-object reconstruction.

\noindent \textbf{3D Hand Pose Estimation.}
Estimating 3D hand pose from RGB images has been proposed for both free hands and hands in-interactions.
FrankMocap~\cite{RongFrankMocap:Integration} is commonly used CNN-based method and has been integrated into many hand-object reconstruction methods~\cite{Hasson2021TowardsVideos, CaoReconstructingWild, Patel2022LearningVideos, YeWhatsHands, ye2023vhoi}.
METRO~\cite{lin2021end-to-end} proposes to use a transformer on top of the CNN feature for regression.
\HAMER~\cite{pavlakos2024reconstructing} is a recent transformer-based method with improved training data scale and higher network capacity.
A concurrent work, WildHands~\cite{Prakash2023Hands}, addresses in-the-wild egocentric hand pose estimation by considering perspective distortion and  trains on relevant in-the-wild data.
In this work, we use \HAMER \cite{pavlakos2024reconstructing} due to its superior performance. We also show \HAMER~leads to better stable grasp reconstruction than FrankMocap in ablations.

\noindent \textbf{3D Hand-Object-Reconstruction.}
Methods are grouped into two categories.

The first category, known-CAD methods, assumes that object CAD models are given and fits 3D shapes into 2D observations. 
These can further be classified into 
learning-based~\cite{TekinH+O:Interactions, LiuSemi-SupervisedTime, Yang2022Artiboost,Wang2023Interacting, Tse2022Collaborative, Yang2020CPF:Interaction, Aboukhadra2023Thor} or
optimisation-based~\cite{Hasson2021TowardsVideos, CaoReconstructingWild, Patel2022LearningVideos} methods.
Learning-based methods learn to jointly reconstruct hands and objects from seen object examples, 
whereas optimisation-based methods address the reconstruction by directly fitting to 2D signals.
RHO~\cite{CaoReconstructingWild} is the first optimisation based single-frame method.
The optimisation-based methods~\cite{CaoReconstructingWild, Hasson2021TowardsVideos, Patel2022LearningVideos} share the same pipeline where hand/object is first independently optimised, followed by joint optimisation with physical terms.
In particular, HOMan~\cite{Hasson2021TowardsVideos} is a generalisation of the single-frame method that incorporates temporal knowledge through the smoothness of mesh vertices over time and the temporal propagation of object pose initialisation.

The second category, CAD-agnostic methods, aims to estimate the hand and object poses without using explicit CAD models. 
Many CAD-agnostic methods~\cite{ye2023vhoi,HassonLearningObjects,YeWhatsHands, Karunratanakul2020GraspingGrasps,Chen2022AlignSDF:Reconstruction} learn object shape priors from the data. 
The other line of methods~\cite{Huang2022ReconstructingVideo,hampali2023inhand,fan2024hold} uses neural networks to fit the underlying object shape from multiple views. However, these methods typically require careful scanning of a held object, making them impractical for use in in-the-wild videos.

Our dataset and method belong to the first category, which is a simplified assumption needed for the challenges of in-the-wild reconstruction. 
Different from above optimisation-based methods, 
our approach examines the object's relative motion and proposes to constrain the relative motion during stable grasp.

\section{Stable Grasp Reconstruction: Problem and Method}
\label{sec:problem_def}

We first define the standard task of hand-object reconstruction, then detail the adjustments we do to focus on stable grasps.
 
The general task of Hand-Object Reconstruction is defined as: given a sequence of $N$ images of hand-object interactions, annotated with hand-side (left v.s. right) and a known object category, the task aims to produce, for every frame $n$, a pair of 3D meshes of the hand and the object w.r.t. the camera. 

Different from the general task, we focus on a special temporal segment of object interactions -- the \textbf{stable grasp} ---which we will define in Sec~\ref{sec:grasp_boundary_def}.
Given a start-end segment of a stable grasp, we aim to produce \textbf{consistent} hand object reconstructions across all frames within the stable grasp.
We refer to this special-case task as Hand-Object Stable Grasp Reconstruction (HO-SGR).

\subsection{Background and Notations}

Following prior works~\cite{Hasson2021TowardsVideos,CaoReconstructingWild,Patel2022LearningVideos}, we use MANO~\cite{Romero2017EmbodiedTogether} to represent the \textbf{hand mesh}, 
which takes as input a finger articulation vector $\theta \in \mathbb{R}^{45}$ 
and outputs the hand mesh with vertices $V_{h} \in \mathbb{R}^{778 \times 3}$ in the \textit{hand coordinate system}. 
We use the recent trained transformer model, \HAMER~\cite{pavlakos2024reconstructing}, to obtain the finger articulations $\Indexed{\theta}$ from individual frames. 
We thus have per-frame hand vertices 
$\Indexed{V_{h}} = \mathrm{MANO}(\Indexed{\theta})$. 
Additionally, the hand-to-camera (\textit{h2c}) pose $\Indexed{T_{h2c}} \in SE(3) $\footnote{$SE(3)$ denotes the space of all combinations of rotations and translations~\cite{MaInvitationTo3DVision}.}, which is defined as the hand wrist orientation and position, is produced by \HAMER~by default and are used to transform $\Indexed{V_{h}}$ to $\Indexed{V_{h:c}}$ in the \textit{camera coordinate system} for each frame.

For the \textbf{object mesh}, we use $V_o \in \mathbb{R}^{|V_o| \times 3}$ to denote the known object vertices in the \textit{object coordinate system}. 
Different from~\cite{Hasson2021TowardsVideos, Patel2022LearningVideos, fan2023arctic}, which estimates the object w.r.t. camera, 
we estimate the object-to-hand~($o2h$) poses $\Indexed{T_{o2h}} \in SE(3)$ and the scalar scale $s \in \mathbb{R}$, which transforms object vertices to $\Indexed{V_{o:h}}$ in the \textit{hand coordinate system} for each frame.
We then use the hand-to-camera~(\textit{h2c}) pose $\Indexed{T_{h2c}}$ to transform $\Indexed{V_{o:h}}$ to $\Indexed{V_{o:c}}$ in the \textit{camera coordinate system} for each frame. 
\begin{equation}
\Indexed{V_{o:c}} = \Indexed{T_{h2c}} ( \Indexed{T_{o2h}} (s * V_o) )
\end{equation}

\subsection{What is a \textit{stable grasp (SG)}?}
\label{sec:grasp_boundary_def}

\begin{figure}[t]
    \centering
    \includegraphics[width=0.7\linewidth]{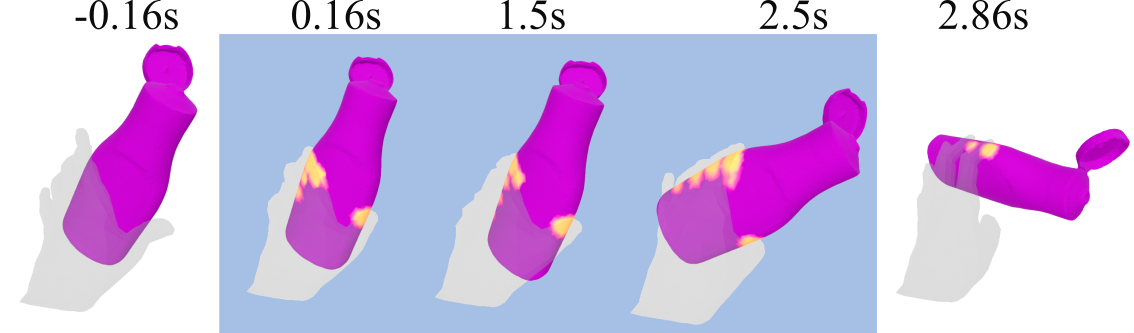}
    \caption{Sample hand-object mesh sequence from ARCTIC. 
    Contact areas (in shiny yellow) are similar within the stable grasp (blue background).
    In -0.16s the hand has no contact with the object.
    }
    \label{fig:visual_grasp_def}
\end{figure}

\noindent \textbf{SG Definition.} The term \textit{stable grasp} has been previously used in human grasp analysis~\cite{Bullock2013HandCentric, Cutkosky1989OnGraspChoice, Feix2016TheTypes}.
While definitions vary, they centre around the object being ``held securely with one hand, irrespective of the hand orientation''~\cite{Feix2016TheTypes}.  
Intuitively, this definition implies the hand maintains a stable contact (area) with the object -- i.e. the same vertices of objects and hands are in contact for the duration of the grasp.
For example, if a hand pours liquid out of a bottle, the hand indeed maintains contact with the same vertices in the bottle throughout manipulation, tilting to pour, then tilting back again upright before putting down. 
While the hand orientation and finger poses change during this interaction, the key here is the consistent stable contact area, which can be quantitatively
computed from hand-object meshes.
We visualise this example in~\cref{fig:visual_grasp_def}.

Formally, for any pair of frames $i$ and $j$ within the stable grasp, 
we use $S_i$ and $S_j$ to denote the in-contact area on the object surface,
and intersection-over-union $\mathrm{IOU}(S_i, S_j)$ between in-contact areas.
Following above intuition, the duration of the stable grasp is defined as
\begin{equation}
\label{eq:auto_grasp_algo}
    \left[l^*, r^*\right] = \underset{l, r}{\mathrm{argmax}} \left(r - l \right) \quad
    \textrm{s.t.} \; \mathrm{IOU}(S_i, S_j) > \tau \;\; \forall l \leq i < j \leq r 
\end{equation}
where $\tau$ specifies the minimum IOU threshold. 
The $argmax (r-l)$ implies the longest duration representing the stable grasp, from its initiation to conclusion. 

\begin{figure}[t]
    \centering
    \includegraphics[width=0.8\textwidth]{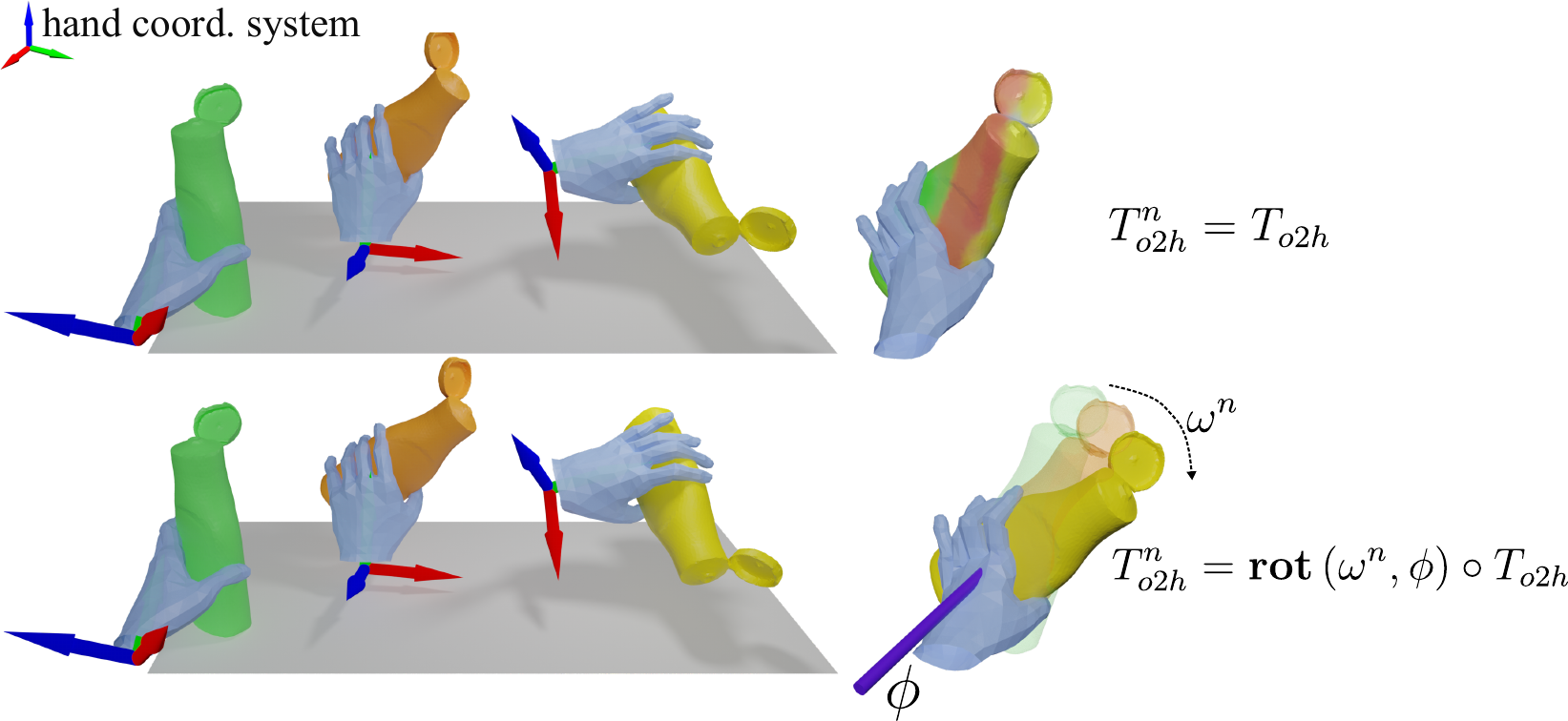}
    \caption{
    To study the object's relative pose, 
    we align the hand coordinate systems (right).
    \textbf{Top}: Made-up sequence with static relative pose -- object is perfectly aligned (mixture of 3 colours).
    \textbf{Bottom}: True sequence showcasing object's motion relative to the hand can be approximated as 1-DoF rotation around axis $\phi$, shown in purple.
    }
    \vspace*{-12pt}
    \label{fig:illustrate_relative_1dof}
\end{figure}

\noindent \textbf{SG Object Motion.} It is critical to note that during the stable grasp, the hand still has dexterity to move or rotate an object relative to the hand pose; 
in other words, the object has non-static motions w.r.t. the hand. 
% \cref{fig:quan_grasp_approx}~(right) gives one such example.
We visualise this distinction in \cref{fig:illustrate_relative_1dof} where we plot one manipulation sequence using two assumptions. The top assumes the object remains static relative to the hand pose -- i.e. when the hand coordinate systems are aligned over the sequence, the object is perfectly aligned.
This top example is actually a made-up one to explain the concept.
The bottom example is a true sequence from the ARCTIC~\cite{fan2023arctic} dataset. 
We showcase that when the hand coordinate system is aligned over time, the object is visibly moving relative to the hand. 
The figure also shows that this motion is not completely free, but is restricted to rotation around a latent axis~$\phi$.
This finding about the object's motion during the stable grasp is one of our paper's contributions.
We detail next the study we carried out, which resulted in this finding.

\noindent \textbf{SG Study.} We perform a study on the 3D-ground-truth dataset ARCTIC~\cite{fan2023arctic} to analyse different quantitative measures within/outside the temporal extent of the stable grasp.
We use the threshold $\tau=0.5$ (Eq~\ref{eq:auto_grasp_algo}) and automatically extract 1303 stable grasp sequences from a variety of objects and subjects throughout dataset (see appendix for details).

\begin{figure}[t]
    \centering
    \includegraphics[width=0.8\textwidth]{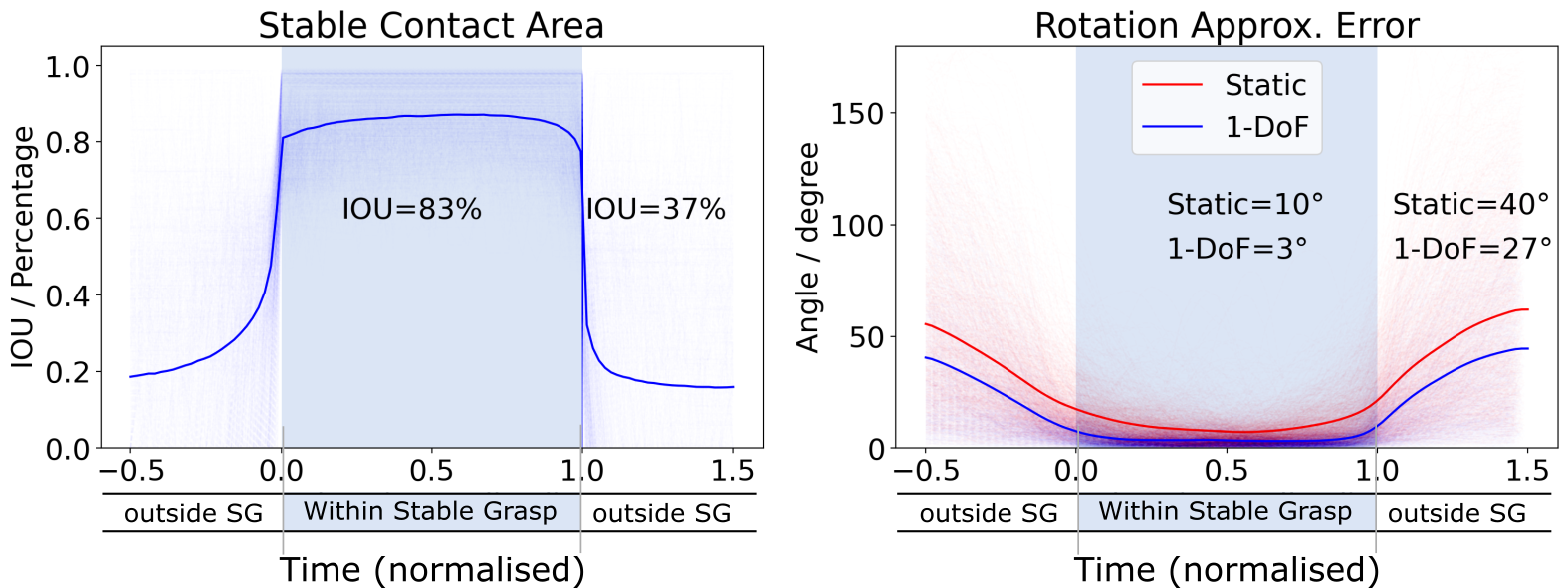}
    \vspace*{-6pt}
    \caption{
    We compare within/outside grasps, analysing object in-contact area (left) and 
    corresponding rotation errors of the static and 1-DoF rotation approximations (right),
    normalising all stable grasp duration for direct comparison (0 to 1 marked with blue background).
    While both the Static and 1-DoF assumptions result in low approximation error within stable grasps, the error of 1-DoF assumption is marginal (right).
    }
    \vspace*{-12pt}
    \label{fig:quan_grasp_approx}
\end{figure}

We present the main finding in Fig~\ref{fig:quan_grasp_approx}.
As anticipated, we show that the in-contact IOU,  drops sharply outside the stable grasp -- plotted in Fig~\ref{fig:quan_grasp_approx}~(left).
That is the contact area remains stable only within the stable grasp's temporal extent.
We similarly compare other quantities (e.g. hand pose changes, finger-tip pose variations and object rotation) within/outside stable grasps (details in appendix).
When we assess the \textbf{relative} object motion to the hand coordinate system within/outside the stable grasp, we note the findings first presented in Fig~\ref{fig:illustrate_relative_1dof}.
Formally, we define a rotation axis $\phi$ around which the object can rotate -- we detail how we find this axis later. If an object is only allowed to rotate around this axis, the motion is restricted from its free 6-DoF to a single rotation angle around this axis -- we thus refer to this as a 1-DoF motion.
The object pose w.r.t. the hand would then be described as
\begin{equation}
\label{eq:1dof}
    \Indexed{T_{o2h}} = \mathbf{rot}\left(\omega^n, \phi\right) \circ T_{o2h}
\end{equation}
i.e., we first apply one global object-to-hand transform $T_{o2h}$ for all frames,
followed by applying the per-frame rotation $\mathbf{rot}\left(\omega^n, \phi\right)$ -- the rotation of angle $\Indexed{\omega}$ around the given rotation axis $\phi$.
Here $\circ$ denotes composition of transformations.

Given the ground truth hand and object poses across a sequence, we can find the optimal rotation axis $\phi$ by minimising the error between the predicted object-to-hand transformation and the ground truth ones, that is:
\begin{equation}
    \underset{\phi, \{\Indexed{\omega}\}, T_{o2h}}{\mathrm{argmin}} \frac{1}{N} \sum_{n=1}^{N} 
    \left\| T_{o2h}^n - T_{gt}^n \right\|^2
\end{equation}
where $T_{gt}^n$ is the ground truth object-to-hand transformation. 
In~\cref{fig:quan_grasp_approx} (right) we plot this within/outside the temporal extent of stable grasp and compare that to the assumption that objects remain static (i.e. does not move relative to the hand). 
We calculate two rotation errors deviated from the ground truth -- the first assumes the object remains static, and the second allows the object to move freely around $\phi$ and calculates the angle error off this axis.
We show that the angle error when using the static assumption to be non-negligible (avg $10^\circ$).
When we use the 1-DoF assumption, the error off the rotation axis $\phi$ is generally low (avg $3^\circ$). 

In appendix, we further confirm these findings on the HOI4D~\cite{Liu2022HOI4D:Interaction} dataset.
This finding on the object's motion within the SG forms the base of our proposed optimisation for HO-SGR, which we present next.

\subsection{Reconstructing Object Poses in a Stable Grasp}
\label{sec:recon_in_grasp}

\begin{figure}[t]
\centering
\includegraphics[width=\linewidth]{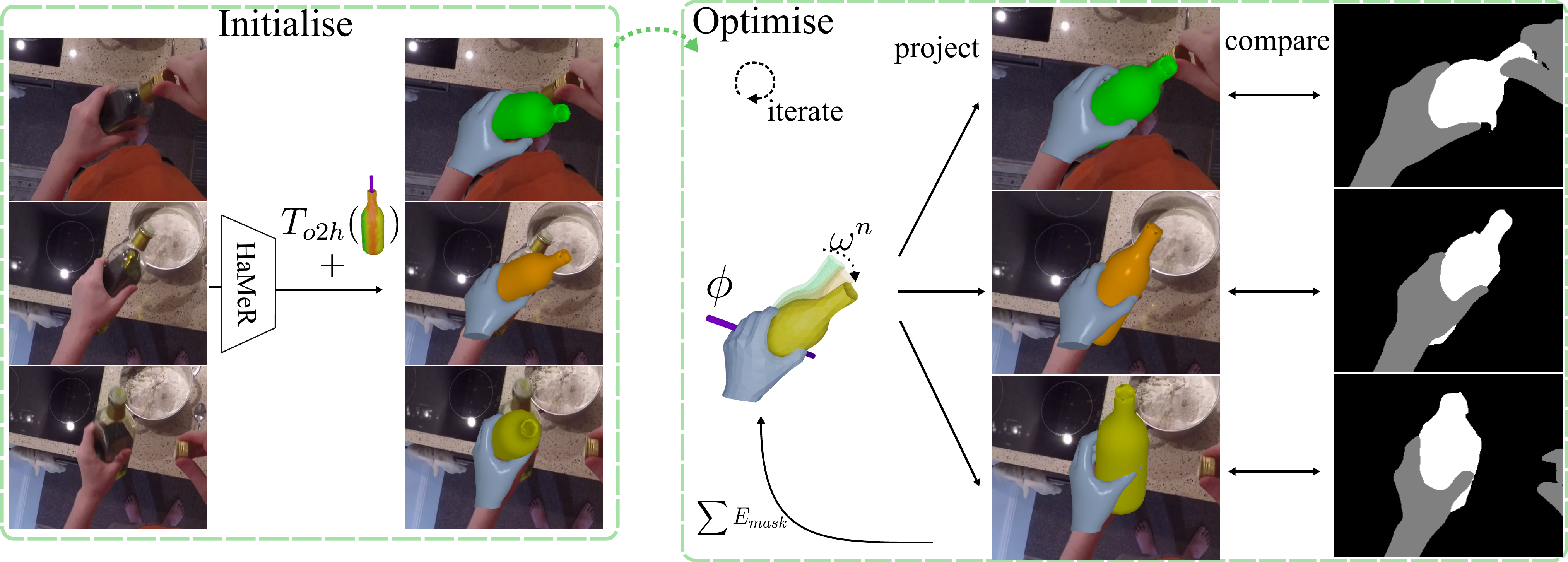}
\caption{
Our proposed reconstruction method. 
We show 3 frames within a stable grasp.
\HAMER~\cite{pavlakos2024reconstructing} produces the hand meshes (rendered in blue) from RGB,
and we set the object-to-hand pose $\Indexed{T_{o2h}}$ to the same $T_{o2h}$ initially.
Then, during each iteration of the optimisation, 
the object's relative pose is optimised to 1-DoF and projected back to individual frames. 
These are compared with ground truth segmentation (right), jointly optimise for all frames. 
We ignore mask computation in hand occluded region (grey in the right figure). 
The physical terms are omitted in this figure.
}
\label{fig:method}
\end{figure}

Recall that we aim to reconstruct the $N$ pairs of 3D meshes of the hand and the object during the stable grasps. 
We use \HAMER~\cite{pavlakos2024reconstructing} to reconstruct the hand pose individually in every frame.
\HAMER~is a fully transformer-based architecture of high accuracy and robustness, trained on very large scale hand data.
Given an RGB image, \HAMER~takes as input the hand detection bounding box with a known hand side, and outputs the MANO finger articulations $\theta^n$ and the hand-to-camera pose $\Indexed{T_{h2c}}$.

We then reconstruct the object relative pose with the
\textit{render-and-compare} approach, 
overviewed in~\cref{fig:method}. 
We propose to leverage our finding on the stable grasp, to optimise the object pose relative to the hand.
It is non-trivial to directly incorporate the stable contact area into the optimisation, as the contact area is not known in advance. 
Instead, we use our finding on the object relative motion, and minimise this to a single degree-of-freedom (1-DoF), i.e. the object rotates along a fixed latent axis to-be-optimised.
As shown in~\cref{sec:grasp_boundary_def}, the 1-DoF object relative pose $\Indexed{T_{o2h}}$ at frame $n$ is given by per-frame rotation $\Indexed{\omega}$, the rotation axis $\phi$ and the global base object-to-hand pose $T_{o2h}$ that takes the object into the hand.

Our main objective function is given by:
\begin{equation}
    \label{eq:totError}
    % \resizebox{\linewidth}{!}{$
    E(\underbrace{\phi, \{\Indexed{\omega}\}, T_{o2h}}_{\Setof{T_{o2h}}}, s; 
    \Setof{\theta}, \Setof{T_{h2c}} ) = 
    \sum\limits_{n=1}^N  \lambda_1 \Indexed{E_{mask}} + \lambda_2\Indexed{E_{push}} + \lambda_3\Indexed{E_{pull}}
    % $}    
\end{equation}
where $\Setof{\theta}$ and $\Setof{T_{h2c}}$  are sets of outputs from \HAMER~and are kept fixed, and $s$ is the scalar object scale to be optimised.

We initialise
the latent axis $\phi$ to the object's z-axis from the CAD model; $\{\Indexed{\omega}\}$ are initialised to zero.  We note the initialisations for $T_{o2h}$ in implementation details.
We then jointly optimise all parameters across frames, 
in particular the rotation axis $\phi$ and the per-frame rotation angles $\Setof{\omega}$.

We use three terms in the optimisation: $\Indexed{E_{mask}}, \Indexed{E_{push}}, \Indexed{E_{pull}}$. We describe these next.
The main term $E_{mask}$ focuses on estimating a reconstruction that best matches the 2D projections of the object masks throughout the sequence. 
We measure the error via sum of pixel differences: 
\begin{equation}
\label{eq:e_mask}
    \Indexed{E_{mask}} = | \Indexed{\mathcal{C}_o} \otimes (\Indexed{\mathcal{M}_o} - {\Pi(\Indexed{V_{o:c}})})|^2_2
\end{equation}
where $\Indexed{\mathcal{M}_o}$ is the object mask which we use for supervision and $\Pi(\cdot)$ is the differentiable projection function~\cite{KatoNeuralRenderer}.
$\Indexed{\mathcal{C}_o}$ is the occlusion-aware mask as in~\cite{Hasson2021TowardsVideos, ZhangPerceivingWild} 
which only computes the error within regions of the object that are not occluded by the hand,
set to 1 for the object and the background, and 0 for the hand.
This masking is critical to avoid penalising the missing parts of the object mask due to in-hand occlusion.

We also employ two additional terms, used in previous works~\cite{Yang2020CPF:Interaction,CaoReconstructingWild,Hasson2021TowardsVideos,Patel2022LearningVideos}. 
We use the physical heuristics $E_{push}$, which pushes the object out of the penetrating region against the hand and a balancing loss $E_{pull}$ which pulls the object to touch these contact regions. 
For exact calculations of these previously used physical heuristics terms, refer to the appendix.

\section{EPIC-Grasps Annotations}
\label{sec:dataset}

With the definition of stable grasp in Sec~\ref{sec:grasp_boundary_def}, 
we now use this to annotate the temporal extents of stable grasps in unscripted egocentric recordings capturing in-the-wild hand-object stable grasps.
This offers a dataset distinct from prior works, where frame-level labels \cite{YangOakInk:Interaction, HassonLearningObjects, CaoReconstructingWild}, 3D supervision \cite{BrahmbhattContactPose:Pose, TaheriGRAB:Objects, Zhang2021ManipNet:Representation} or recordings specifically collected to evaluate grasps with no underlying action \cite{HampaliHOnnotate:Poses, KwonH2O:Recognition, ChaoDexYCB:Objects} are typically used.
Instead, we aim to find stable grasps from within egocentric videos recorded of daily actions in an unscripted manner.

We next detail our pipeline to annotate EPIC-Grasps:

\noindent \textbf{1. Identifying candidate clips to annotate.} The ultimate goal of hand-object reconstruction is to operate for any rigid or dynamic objects, including novel classes. 
However, current approaches for reconstruction of unknown objects are still in their infancy \cite{YeWhatsHands, ye2023vhoi,Sucar2020NodeSLAM:Reconstruction,Huang2022ReconstructingVideo,swamy2023showme} (we show failure cases in appendix). 
We thus restrict our first-attempt at reconstructing stable grasps in-the-wild to objects of a known category.
Note that this is distinct from instance CAD models -- the general CAD model of a bottle might not exactly match all bottles in daily life.
We exclude very small objects and shortlist 9 object categories frequently used in kitchens: plate, bowl, bottle, cup, mug, can, pan, saucepan, glass\footnote{For the \textbf{object mesh}, we made one category CAD model in Blender~\cite{blender}.}.

We begin with the frame-level hand-object interaction annotations from VISOR~\cite{VISOR2022}, in which we identified clips where the a hand is in contact with one of the categories above. 
The annotations also inform us which hand is in contact with the object (i.e. left or right).
We then use the fine-grained action labels available from EPIC-KITCHENS~\cite{Damen2022RESCALING, DamenScalingDataset} to locate the initial temporal extent for actions involving any of these categories. 

\noindent \textbf{2. Annotating Stable Grasps.}
Two annotators were asked to label the start-and-end frames following the stable grasp definition.
 We discard segments when (i)~a segment does not contain any stable grasp (ii)~both the hand and object are out-of-view during the sequence or (iii)~the object does not match the category CAD model specified.

 \noindent \textbf{3. 2D Hand and Object Segmentation Annotations.}
In total, we label 2431 video clips of stable gasps from 141 distinct videos in 31 kitchens~\cite{Damen2022RESCALING}.
For each clip, we provide a start and end time of the stable grasp, as well as 319,661 segmentation masks for the hand and the object during the stable grasp from the dense VISOR annotations~\cite{VISOR2022}.
Of these, 1446 contain left hand stable grasps and 985 contain right hand stable grasps.
Note that the majority of subjects recording the EPIC-KITCHENS dataset are right-handed---which implies the left hand is more frequently used to steady the object during manipulation.
Compared to previous efforts \cite{HampaliHOnnotate:Poses, Liu2022HOI4D:Interaction, KwonH2O:Recognition, ChaoDexYCB:Objects, swamy2023showme}, most clips in EPIC-Grasps contains \textit{both hands in view} manipulating the same or different objects.
We provide the breakdown per category and grasp duration in appendix.

\begin{table*}[t]
\centering
\resizebox{\textwidth}{!}{%
% \begin{tabular}{ l c | c p{1cm} p{1cm} | p{1.2cm} p{1.2cm} p{1.2cm} p{1.2cm} p{1.2cm} | p{1.5cm} p{1.2cm}}
\begin{tabular}{ l r @{\hspace{0.2cm}} | c p{1cm} p{0.8cm} | r r r r @{\hspace{0.4cm}} r @{\hspace{0.2cm}} | p{1.6cm} p{1cm}}
& &\multicolumn{3}{c|}{\textbf{Characteristics}} &\multicolumn{5}{c|}{\textbf{Stats}} &\multicolumn{2}{c}{\textbf{Labels}}\\

Dataset &Year & \small{In-the-wild} & \parbox{1cm}{\small{Funct.}\\ Intent} & \small{Ego} &\#Env & \#Sub & \#Cat & \#Inst & \#Seq & Pose GT& \parbox{3cm}{Stable \\ Grasp} \\
 \hline
  FPHA~\cite{Garcia-HernandoFirst-PersonAnnotations} &2018 & \xmark & \cmark & \cmark & 3 & 6 & 4 & 4 & 1,175 & 3D & \xmark \\

 HO3D~\cite{HampaliHOnnotate:Poses} &2020  & \xmark & \xmark & \xmark & 1 & 10 & 10 & 10 & 65 & 3D & \cmark (part)  \\
 ContactPose~\cite{BrahmbhattContactPose:Pose} &2020 & \xmark & \cmark & \xmark & 1 & 50 & 25 & 25 & 2,306 & 3D & \cmark \\
 GRAB~\cite{TaheriGRAB:Objects} &2020 & \xmark & \cmark & \xmark & 1 & 10 & 51 & 51 & 1,334 & 3D & \xmark \\
 
 H2O~\cite{KwonH2O:Recognition} &2021 & \xmark & \cmark & \cmark & 3 & 4 & 8 & 8 & 24 & 3D & \xmark \\
 % For H2O, download label_splits and count how many subject1_h1-like
 DexYCB~\cite{ChaoDexYCB:Objects} &2021 & \xmark & \xmark & \xmark & 1 & 10 &  20 & 20 & 1,000 & 3D & \xmark\\
 
 HOI4D~\cite{Liu2022HOI4D:Interaction} &2022 & \xmark & \cmark & \cmark & 610 & 9 & 20 & 800 & 5,000 & 3D & \xmark \\
 Assembly101~\cite{SenerAssembly101:Activities} &2022 & \xmark & \cmark &\xmark & 1 & 53 & 15 & 15 & 4,321 & 3D Hand$^*$ & \xmark\\
 OakInk~\cite{YangOakInk:Interaction} &2022 & \xmark & \cmark & \xmark & 1 & 12 & 32 & 100 & 1,356 & 3D & \xmark \\
 % For oakInk, download the data and look at how many source.txt they have; also they Table-1 says 1K interactions
  ARCTIC~\cite{fan2023arctic} &2023 & \xmark & \cmark & \cmark & 1 & 9$^\dagger$ & 11 & 11 & 339 & 3D & \xmark \\
 ARCTIC-Grasps (ours) &2024 & \xmark & \cmark & \cmark & 1 & 9 & 11 & 11 & 1,303 & 3D & \cmark \\
 \hline 
% Yale \\
 Core50~\cite{LomonacoCORe50:Recognition} &2017 & \cmark & \xmark & \xmark & 11 & - & 10 & 50 & 550 & 2D Mask  & \xmark \\
 MOW~\cite{CaoReconstructingWild, Patel2022LearningVideos} &2021 & \cmark & \cmark & \xmark & 500 & 500 & 121 & 500 & 500 & \xmark & \xmark\\ 
 EPIC-Grasps (ours) &2024 & \cmark & \cmark & \cmark & 141 & 31 & 9 & $\sim$390 & 2,431 & 2D Mask & \cmark \\
 \hline
\end{tabular}
    }
    \caption{Dataset Comparison: $^*$: cannot be used for hand-object reconstruction as object poses or segments are not provided. $^\dagger$: subjects in the released train/val set.} 
    % ARCTIC: Two subjects are in the held out test set.}
    \vspace*{-12pt}
    \label{tab:datasetCompare}
\end{table*}

~\cref{tab:datasetCompare} compares our dataset with currently available and regularly used datasets for hand-object reconstruction.
In the first section, we show datasets collected with 3D ground truth and add the 1303 stable grasp sequences previously noted in~\cref{sec:grasp_boundary_def}. We refer to these as ARCTIC-Grasps.
We note that previous datasets, including those that requiring 3D ground truth, are limited in the environmental setup to allow for multi-camera views or motion capture tracking.
This limits the appearance diversity and interactions are not functional.
In~\cref{tab:datasetCompare} we also report the functional intent of various dataset (i.e. object is used with intent). 
EPIC-Grasps is the first to capture in-the-wild egocentric videos with functional intent.
Of particular note is the HOI4D~\cite{Liu2022HOI4D:Interaction} where diversity is targeted with a large number of indoor rooms. However, this is a constructed setup, where objects are selected from a given pool.
In EPIC-Grasps, subjects are recording in their personal kitchens and using their chosen objects as part of longer activities.

\section{Experiments}
\label{sec:experiments}

The experiment section is structured as follows:
in~\cref{sec:implementation} and~\cref{sec:define_metrics} we describe the implementation details of our method and the quantitative metrics respectively.~\cref{sec:quan_arctic} and~\cref{sec:quan_epic} report results of our method and baselines on ARCTIC-Grasps and EPIC-Grasps, respectively. Finally~\cref{sec:ablation} conducts ablation studies on our method.

\subsection{Implementation Details}
\label{sec:implementation}

We follow the  convention in HOMan~\cite{Hasson2021TowardsVideos} which sparsely and linearly samples 30 frames from the sequence for optimisation and evaluation.
We use 50 initialisations of object rotation and 1 global translation for each object, from ground truth of ARCTIC to find these initialisations, following~\cite{Hasson2021TowardsVideos}.
The optimisation process takes 1.5 minutes on a 2080Ti for one 30-frame sequence on average.

The error $E_{mask}$ is defined in pixels whilst $E_{push}$ and $E_{pull}$ are defined in 3D space (mm), and are related by the camera's focal length $f$. We introduce a focal scaling factor: ${\lambda_f = f * \texttt{render\_size}}$.
We use $\texttt{render\_size} = 256$ and set $\lambda_1 = 1$, $\lambda_2 = 0.1 * \lambda_f$ and $\lambda_3 = 0.1 * \lambda_f$; here the focal length $f$ is estimated by \HAMER~\cite{pavlakos2024reconstructing}.
Please refer to appendix for further implementation details.

\subsection{Baselines and Quantitative Metrics}
\label{sec:define_metrics}

We focus on baselines that do not require training on the dataset in advance. We thus use 4 baselines to compare to:
\begin{itemize}[leftmargin=5mm,itemsep=-1.5ex,partopsep=1ex,parsep=2ex]
\item \textbf{HOMan}~\cite{Hasson2021TowardsVideos}---a common CAD-based baseline that progressively optimises the object pose relative to the hand. We use the ground truth masks as in our method for fair comparison.
\item \textbf{Single Frame}---We predict the hand and object meshes independently for each frame, without utilising any temporal optimisation or the knowledge of stable grasps.
\item \textbf{Static}---A version of our method where objects are not allowed any motion when within the stable grasp, minimising the overall rotation and translation of the object.
\item \textbf{Dynamic}---A variation of our method where objects are allowed to move freely within 6-DoF.
\end{itemize}
We note that 
RHOV~\cite{Patel2022LearningVideos} does not have released code to be used as a baseline. 

We propose quantitative metrics to measure the correctness of the predicted object poses within the stable grasp.
Given the knowledge that stable grasps maintain a consistent contact area between the hand and the object throughout the sequence, we measure this contact area over time.
However, a reconstruction can maintain a stable contact area but be completely erroneous.
We thus propose to combine a measure of correctness of reconstruction with the stability of the contact area to produce a robust metric. We explain this next.

\noindent \textbf{Average Distance (ADD \%)}. Following~\cite{krull2015learning, Xiang2017PoseCNN, Hodan2020Epos}, we measure the distance of corresponding vertices between GT and predicted object vertices in the hand coordinate system, and average this distance over vertices and frames. ADD is 1 for a sequence if the average distance is less than 10\% of the object's diameter, and 0 otherwise.

\noindent \textbf{Average Stable Contact Area at ADD Success (SCA-ADD \%)}. 
When a pose is considered correct for a sequence, i.e. ADD is 1, we measure the stable contact area across the sequence, defined as the average IOU of in contact area between each pair of frames~(\cref{sec:grasp_boundary_def}). SCA-ADD is set to 0 when ADD is set to 0 (average distance below threshold). We average SCA-ADD over all examples.  

\noindent \textbf{Intersection-over-Union (IOU \%)}. 
We use IOU as a proxy of pose accuracy when 3D GT is not available. We measure the Intersection-over-Union between the ground truth mask and the rendered mask for the object in camera view. We report average IOU across all frames. Due to occlusion with other components in the scene, only the non-occluded area of the rendered projection is used.

\begin{table*}[t]
    \centering
    \resizebox{\textwidth}{!}{%
    \begin{tabular}{l|cccc|cccc|cccc|cccc|cccc}% \begin{tabular}{l|rrrr|rrrr|rrrr|rrrr|rrrr}
\toprule
 & \multicolumn{4}{c|}{Single Frame} & \multicolumn{4}{c|}{HOMan~\cite{Hasson2021TowardsVideos}} & \multicolumn{4}{c|}{Static} & \multicolumn{4}{c|}{Dynamic} & \multicolumn{4}{c}{1-DoF} \\
Category & IOU & SCA-IOU & ADD & SCA-ADD & IOU & SCA-IOU & ADD & SCA-ADD & IOU & SCA-IOU & ADD & SCA-ADD & IOU & SCA-IOU & ADD & SCA-ADD & IOU & SCA-IOU & ADD & SCA-ADD \\
\midrule
box & 93.2 & 17.5 & 9.4 & 3.1 & 83.5 & 30.9 & 33.3 & 16.9 & 90.9 & \cellcolor{green!25}67.3 & 37.7 & \cellcolor{green!25}28.7 & \cellcolor{green!25}95.2 & 41.7 & \cellcolor{green!25}57.2 & \cellcolor{yellow!25}25.4 & \cellcolor{yellow!25}94.2 & \cellcolor{yellow!25}56.7 & \cellcolor{yellow!25}39.1 & 23.6 \\
capsulemachine & 79.3 & 17.8 & 8.4 & 3.3 & 84.0 & 34.4 & 42.1 & 24.5 & 79.8 & 34.9 & 48.4 & \cellcolor{yellow!25}34.1 & \cellcolor{green!25}88.8 & \cellcolor{yellow!25}41.6 & \cellcolor{yellow!25}53.7 & 29.2 & \cellcolor{yellow!25}86.1 & \cellcolor{green!25}51.4 & \cellcolor{green!25}62.1 & \cellcolor{green!25}41.2 \\
espressomachine & 82.8 & 19.1 & 22.8 & 8.3 & 84.2 & 36.9 & 44.6 & 28.8 & 82.6 & \cellcolor{yellow!25}49.2 & 48.5 & \cellcolor{green!25}36.2 & \cellcolor{green!25}89.6 & 43.1 & \cellcolor{green!25}66.3 & 34.5 & \cellcolor{yellow!25}87.4 & \cellcolor{green!25}54.9 & \cellcolor{yellow!25}54.5 & \cellcolor{yellow!25}35.6 \\
ketchup & 81.3 & 29.3 & 30.2 & 16.4 & 72.0 & 18.0 & 15.1 & 9.3 & 75.3 & 32.7 & 51.9 & 37.8 & \cellcolor{green!25}89.4 & \cellcolor{green!25}56.0 & \cellcolor{yellow!25}62.3 & \cellcolor{yellow!25}40.2 & \cellcolor{yellow!25}84.0 & \cellcolor{yellow!25}55.4 & \cellcolor{green!25}64.2 & \cellcolor{green!25}45.0 \\
laptop & 87.8 & 24.9 & 16.7 & 7.0 & 82.1 & 35.3 & 43.8 & 28.0 & 86.0 & \cellcolor{yellow!25}62.0 & 45.1 & \cellcolor{yellow!25}37.1 & \cellcolor{green!25}93.3 & 50.2 & \cellcolor{green!25}63.2 & 34.6 & \cellcolor{yellow!25}91.4 & \cellcolor{green!25}68.0 & \cellcolor{yellow!25}54.9 & \cellcolor{green!25}40.8 \\
microwave & 85.7 & 17.1 & 29.5 & 8.0 & 87.0 & 36.1 & \cellcolor{green!25}56.2 & 27.3 & 85.4 & \cellcolor{yellow!25}51.5 & 50.9 & \cellcolor{green!25}35.3 & \cellcolor{green!25}90.3 & 35.2 & 53.6 & 23.0 & \cellcolor{yellow!25}89.0 & \cellcolor{green!25}52.5 & \cellcolor{yellow!25}55.4 & \cellcolor{yellow!25}34.8 \\
mixer & 79.4 & 12.1 & 14.8 & 4.9 & 85.3 & 34.3 & 45.1 & 26.7 & 79.3 & 37.9 & 48.4 & \cellcolor{yellow!25}32.2 & \cellcolor{green!25}89.1 & \cellcolor{yellow!25}38.4 & \cellcolor{yellow!25}59.0 & 29.0 & \cellcolor{yellow!25}87.0 & \cellcolor{green!25}52.0 & \cellcolor{green!25}62.3 & \cellcolor{green!25}39.6 \\
notebook & 87.2 & 25.0 & 6.6 & 2.8 & 85.7 & 38.7 & 33.8 & 20.8 & 84.5 & \cellcolor{yellow!25}57.8 & 43.0 & 33.2 & \cellcolor{green!25}93.5 & 52.8 & \cellcolor{green!25}63.6 & \cellcolor{yellow!25}35.0 & \cellcolor{yellow!25}90.9 & \cellcolor{green!25}65.5 & \cellcolor{yellow!25}55.6 & \cellcolor{green!25}38.7 \\
phone & 82.9 & 25.7 & 15.1 & 6.8 & 80.9 & 39.5 & 28.1 & 19.9 & 77.8 & 36.7 & \cellcolor{yellow!25}39.7 & \cellcolor{yellow!25}29.2 & \cellcolor{green!25}91.1 & \cellcolor{yellow!25}58.8 & 37.7 & 23.5 & \cellcolor{yellow!25}88.1 & \cellcolor{green!25}61.9 & \cellcolor{green!25}54.1 & \cellcolor{green!25}37.6 \\
scissors & 48.5 & 0.0 & 14.0 & 9.3 & 40.9 & \cellcolor{yellow!25}5.2 & 7.0 & 5.2 & 48.2 & 0.0 & 47.4 & 36.3 & \cellcolor{green!25}73.9 & \cellcolor{green!25}18.2 & \cellcolor{yellow!25}56.1 & \cellcolor{yellow!25}38.9 & \cellcolor{yellow!25}62.4 & 2.7 & \cellcolor{green!25}68.4 & \cellcolor{green!25}51.8 \\
waffleiron & 85.7 & 13.1 & 3.1 & 0.9 & 86.5 & 36.3 & 45.0 & 26.2 & 84.7 & \cellcolor{yellow!25}48.3 & 59.5 & \cellcolor{yellow!25}39.2 & \cellcolor{green!25}92.7 & 44.3 & \cellcolor{green!25}81.7 & 38.1 & \cellcolor{yellow!25}91.2 & \cellcolor{green!25}55.9 & \cellcolor{yellow!25}69.5 & \cellcolor{green!25}43.7 \\
\hline
All & 83.3 & 19.5 & 15.0 & 6.0 & 81.4 & 33.1 & 37.1 & 22.0 & 81.4 & \cellcolor{yellow!25}46.6 & 46.9 & \cellcolor{yellow!25}34.2 & \cellcolor{green!25}\textbf{90.8} & 45.5 & \cellcolor{green!25}\textbf{59.6} & 31.5 & \cellcolor{yellow!25}88.1 & \cellcolor{green!25}\textbf{55.7} & \cellcolor{yellow!25}57.3 & \cellcolor{green!25}\textbf{38.5} \\
\bottomrule
\end{tabular}
    }
    \caption{Results on ARCTIC-Grasps. 
    \colorbox{green!25}{Green} shows the \colorbox{green!25}{best} performing method per metric and \colorbox{yellow!25}{yellow} shows the \colorbox{yellow!25}{second} best.}
\vspace*{-12pt}
    \label{tab:arctic_results}
\end{table*}

\begin{figure}[t]
    \centering
    \includegraphics[width=\linewidth]{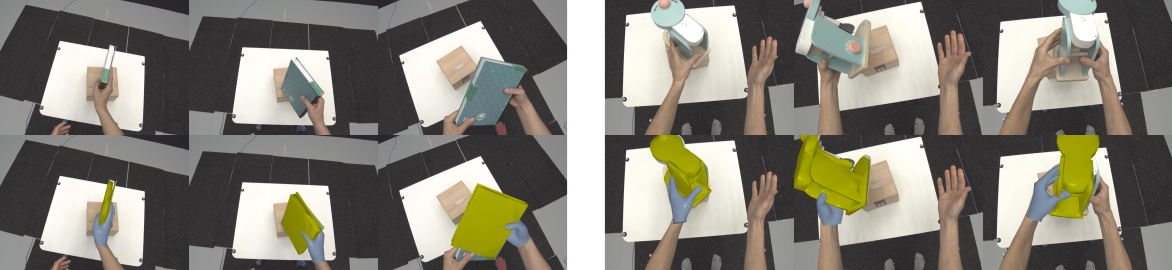}
    \caption{Two stable grasp reconstructions by 1-DoF method on ARCTIC-Grasps.
    }
    \vspace*{-12pt}
    \label{fig:qual_arctic}
\end{figure}

\noindent \textbf{Average Stable Contact Area at high IOU (SCA-IOU \%)}. 
Using IOU as proxy for correct reconstructions, we analogously report SCA when IOU is more than certain thresholds. We use 80\% as threshold on ARCTIC-Grasps and report both 80\% and 60\% on EPIC-Grasps. SCA-IOU is set to 0 when IOU is below the threshold. We report average SCA-IOU.

The first two metrics, ADD and SCA-ADD, are evaluated when 3D ground truths are available---e.g. ARCTIC-Grasps dataset; and the other two are the corresponding 2D proxy, which we use for EPIC-Grasps.

\subsection{Results on ARCTIC-Grasps}
\label{sec:quan_arctic}

\cref{tab:arctic_results} compares results on ARCTIC-Grasps, using all the metrics. 
We here focus on the object metrics, using ground truth hand mesh and a known object scale. 
Whilst Dynamic and 1-DoF both have significantly higher ADD results, 1-DoF maintains the best SCA when grasps are successfully reconstructed (high SCA-ADD).
Static has considerably high SCA, but
IOU and ADD are significantly lower than 1-DoF in every category.
HOMan only performs best on the ``microwave'' object sequences.
Single Frame performs worst in every case. 
We show qualitative examples in~\cref{fig:qual_arctic}.

\begin{table}[t]
    \centering
    \resizebox{0.8\linewidth}{!}{%
    \begin{tabular}{l|ccc|ccc|ccc}
\toprule
 & \multicolumn{3}{c|}{HOMan~\cite{Hasson2021TowardsVideos}} & \multicolumn{3}{c|}{Dynamic} & \multicolumn{3}{c}{1-DoF} \\
Category & IOU & SCA@0.8 & SCA@0.6 & IOU & SCA@0.8 & SCA@0.6 & IOU & SCA@0.8 & SCA@0.6 \\
\midrule
bottle & 56.7 & 3.4 & 5.8 & \cellcolor{green!25}75.5 & \cellcolor{yellow!25}21.8 & \cellcolor{yellow!25}36.6 & \cellcolor{yellow!25}72.1 & \cellcolor{green!25}26.1 & \cellcolor{green!25}51.3 \\
bowl & 54.4 & 1.1 & 1.6 & \cellcolor{green!25}58.0 & \cellcolor{yellow!25}9.3 & \cellcolor{yellow!25}19.8 & \cellcolor{yellow!25}56.2 & \cellcolor{green!25}11.1 & \cellcolor{green!25}25.7 \\
can & 48.3 & 3.4 & 5.1 & \cellcolor{green!25}56.0 & \cellcolor{yellow!25}13.7 & \cellcolor{yellow!25}20.1 & \cellcolor{yellow!25}54.1 & \cellcolor{green!25}16.5 & \cellcolor{green!25}26.6 \\
cup & 56.9 & 5.5 & 7.8 & \cellcolor{green!25}67.5 & \cellcolor{yellow!25}16.5 & \cellcolor{yellow!25}38.9 & \cellcolor{yellow!25}65.9 & \cellcolor{green!25}18.4 & \cellcolor{green!25}46.1 \\
glass & 55.4 & 3.0 & 4.2 & \cellcolor{green!25}65.2 & \cellcolor{yellow!25}14.8 & \cellcolor{yellow!25}30.1 & \cellcolor{yellow!25}62.5 & \cellcolor{green!25}15.7 & \cellcolor{green!25}37.8 \\
mug & 59.4 & 3.9 & 6.6 & \cellcolor{green!25}63.8 & \cellcolor{yellow!25}6.3 & \cellcolor{yellow!25}29.6 & \cellcolor{yellow!25}62.6 & \cellcolor{green!25}10.8 & \cellcolor{green!25}38.6 \\
pan & 48.3 & 0.5 & 1.9 & \cellcolor{yellow!25}48.9 & \cellcolor{yellow!25}4.3 & \cellcolor{yellow!25}12.8 & \cellcolor{green!25}49.1 & \cellcolor{green!25}7.0 & \cellcolor{green!25}20.4 \\
plate & 61.1 & 1.0 & 1.6 & \cellcolor{green!25}68.1 & \cellcolor{green!25}17.6 & \cellcolor{yellow!25}30.1 & \cellcolor{yellow!25}65.2 & \cellcolor{yellow!25}16.3 & \cellcolor{green!25}34.6 \\
saucepan & 51.1 & 0.4 & 2.4 & \cellcolor{green!25}57.5 & \cellcolor{yellow!25}5.4 & \cellcolor{yellow!25}25.7 & \cellcolor{yellow!25}56.8 & \cellcolor{green!25}8.5 & \cellcolor{green!25}34.6 \\
\hline
All & 54.9 & 1.9 & 3.3 & \cellcolor{green!25}\textbf{61.7} & \cellcolor{yellow!25}12.2 & \cellcolor{yellow!25}25.3 & \cellcolor{yellow!25}59.9 & \cellcolor{green!25}\textbf{14.1} & \cellcolor{green!25}\textbf{32.9} \\
\bottomrule
\end{tabular}
     }
    \caption{Results on EPIC-Grasps.
    \colorbox{green!25}{Green} for \colorbox{green!25}{best} and \colorbox{yellow!25}{yellow} shows the \colorbox{yellow!25}{second} best.}
    \vspace*{-12pt}
    \label{tab:epic_results}
\end{table}

\begin{figure}[t]
    \centering
    \includegraphics[width=0.8\linewidth]{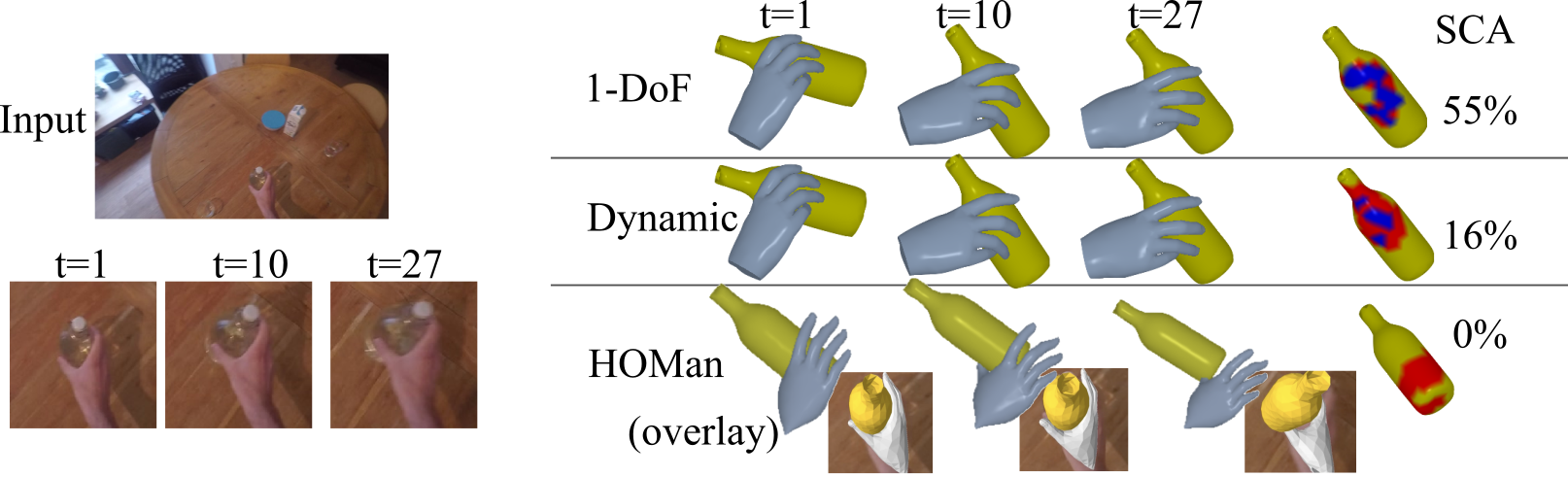}
    \caption{Qualitative comparison between 1-DoF, Dynamic and  HOMan~\cite{Hasson2021TowardsVideos} on EPIC-Grasps. We show input frames, reconstruction results and the corresponding SCA.
    }
    \vspace*{-12pt}
    \label{fig:qual_compare}
\end{figure}

\subsection{Results on EPIC-Grasps}
\label{sec:quan_epic}

\cref{tab:epic_results} compares results on EPIC-Grasps dataset 
using proxy metrics IOU and SCA-IOU at two thresholds (0.8 and 0.6). 
We compare 1-DoF to the best variation from ARCTIC-Grasps: Dynamic. 
1-DoF achieves the best SCA-IOU metric for both thresholds for every object category and overall.

\cref{fig:qual_compare} shows comparative results. 
Despite acceptable overlay, HOMan does not maintain the stable grasp and the object is in fact sliding in the hand over time.
The Dynamic baseline has significantly lower SCA with the contact area drifting over time.
Our proposed 1-DoF method is able to reconstruct the stable grasp as it benefits from the constrained relative pose assumption.

\begin{figure*}[t]
\centering
\includegraphics[width=\textwidth]{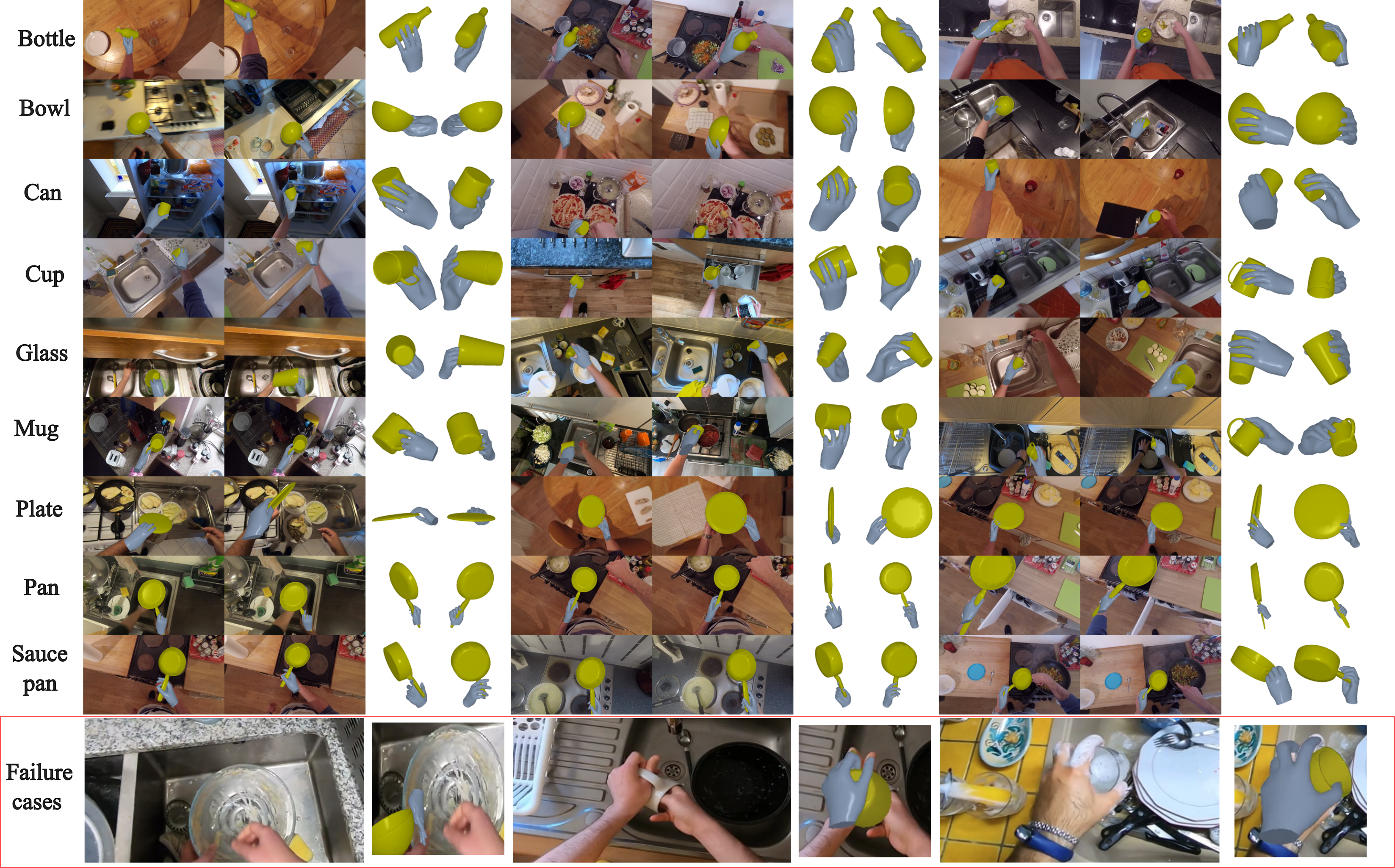}
\caption{Qualitative Results on EPIC-Grasps (3 examples/ category); projected reconstruction results and reconstruction in rotated views.
Bottom: failure cases due to wrong hand pose (left), extreme occlusion (mid) and limited views (right).
}
\vspace*{-12pt}
    \label{fig:bigqual}
\end{figure*}

\cref{fig:bigqual} shows reconstruction results of our 1-DoF approach on the EPIC-Grasps dataset. We showcase three examples per category. From the images, the challenge in EPIC-Grasps is evident with the hand and object occupying a small proportion of the image. In most examples, the palm and fingers are concealed. 
Importantly, all grasps are functional -- e.g. picking a cup from the drainer.

\subsection{Ablation Studies}
\label{sec:ablation}

%\afterpage{
%\clearpage}

In this section, we ablate the hand pose estimator and loss functions under the in-the-wild setting, on full EPIC-Grasps dataset.
The 1-DoF assumption of the method has been ablated against other variants in~\cref{tab:arctic_results}. 

\noindent \textbf{Ablation on the hand pose estimation method.} 
In~\cref{tab:ablate_frank_vs_hamer}, 
we compare \HAMER~against FrankMocap~\cite{RongFrankMocap:Integration},
a commonly used hand predictor in other works~\cite{Hasson2021TowardsVideos, CaoReconstructingWild,Patel2022LearningVideos}.
\HAMER~consistently outperforms the results with FrankMocap on all metrics,
which is in line with the higher capacity of \HAMER~over FrankMocap. We include a detailed table in appendix.

\begin{table}[t]
    \centering
    % \resizebox{\textwidth}{!}{%
    \begin{tabular}{l|ccc}
    \toprule
    Method & IOU & SCA@0.8 & SCA@0.6 \\
    \hline
    1-DoF + FrankMocap~\cite{RongFrankMocap:Integration}  & 55.1 & 7.2 & 20.3 \\
    1-DoF + HaMeR~\cite{pavlakos2024reconstructing} & 59.9 & 14.1 & 32.9 \\
    \bottomrule
    \end{tabular}
    % }
    \caption{Ablation on the hand pose estimator.}
    \vspace*{-12pt}
    \label{tab:ablate_frank_vs_hamer}
\end{table}

\begin{table}[t]
\begin{minipage}{.5\linewidth}
    \centering
    \begin{tabular}{cc|ccc}
    \toprule
    $\lambda_2 \quad$ & $\lambda_3 \quad$ & IOU & SCA@0.8 & SCA@0.6 \\
    \hline
        0.0 &0.0 & 62.5& 15.1 & 27.5 \\
        \rowcolor{blue!10} 0.1 &0.1 & 59.9 & 14.1 & 32.9 \\
        1.0 &1.0 & 53.4 & 12.2 & 25.3 \\
    \bottomrule
    \end{tabular}
    \caption{Ablation on the physical terms.
    We highlight our choice of $\lambda_2$ and $\lambda_3$ (blue).
    }
    \label{tab:ablate_physical_terms}
\end{minipage}
\hfill
\hspace{0.05\linewidth}
\begin{minipage}{.45\linewidth}
\centering
    \begin{tabular}{c|ccc}
    \toprule
    $\Indexed{\mathcal{C}_o}$ & IOU & SCA@0.8 & SCA@0.6 \\
    \hline
    \xmark  & 51.0 & 3.5 & 20.4 \\
    \cmark & 59.9 & 14.1 & 32.9 \\
    \bottomrule
    \end{tabular}
    \caption{Ablation on occlusion-aware mask $\Indexed{\mathcal{C}_o}$.}
    \label{tab:ablate_occlusion_aware_mask}
\end{minipage}
\end{table}

\noindent \textbf{Ablation on physical terms $E_{push}$ and $E_{pull}$.}
In~\cref{tab:ablate_physical_terms}, we ablate the physical terms $E_{push}$ and $E_{pull}$ introduced in the loss function~\cref{eq:totError}. Note that while removing physical terms ($\lambda_2 = \lambda_3 = 0$) slightly improves IOU and SCA@0.8, it significantly decreases SCA@0.6 by $5.4\%$.

\noindent \textbf{Ablation on Occlusion Mask.}
In~\cref{tab:ablate_occlusion_aware_mask}, 
we show that using $\Indexed{\mathcal{C}_o}$ is critical in the optimisation, 
whereas removing $\Indexed{\mathcal{C}_o}$ drastically reduces the results.

\subsection{Limitations}

Whilst results in-the-wild are very promising, our pipeline relies on \HAMER~as a first stage. Despite the robustness incorporated by the multiple-view joint optimisation, our method fails when the predicted hand poses are wrong (see~\cref{fig:bigqual}). 
Our method also struggle with extreme occlusions and ambiguity from limited views.
The 1-DoF assumption also relies on the knowledge of stable grasp duration, which is liable to fail outside grasps.

Another limitation of our approach is its reliance on the knowledge of the category's CAD model. Current CAD-agnostic methods~\cite{YeWhatsHands} struggle in-the-wild --
we showcase failures in appendix. However, removing the limitation of CAD model knowledge is a clear future direction.

%-------------------------------------------------------------------------

\section{Conclusion}

In this paper we propose the task of hand-object stable grasp reconstruction.
By focusing on stable grasps, we can jointly optimise multiple frames given the knowledge of how objects move relative to the hand within grasp.
We further build a challenging in-the-wild egocentric videos dataset for the task. The dataset contains 2D masks as pseudo ground truths and manual temporal labels of stable grasps. 
We evaluate the method's ability without 3D ground truth by measuring the 2D IOU of object masks along with measuring the stable contact area over time. 
We show that the constrained object motion achieves better hand-object reconstructions within the stable grasp.

By investigation stable grasps, we hope to encourage more works to quantitatively evaluate methods in-the-wild. 

\noindent\textbf{Acknowledgements.}
Research is supported by EPSRC UMPIRE (EP/T004991/1). Z. Zhu is supported by UoB-CSC Scholarship.

\bibliographystyle{splncs04}
\bibliography{zhifan_rebibed}

\clearpage
\appendix
% \twocolumn[
% \centering
% \Large
\noindent \textbf{\Large Appendix}\\
% ] %< twocolumn

This appendix is structured as follows:
\cref{sec:epic_grasp_stats} presents statistics of EPIC-Grasps annotation.
\cref{sec:arctic_grasp_stats} presents statistics of ARCTIC-Grasps annotation.
\cref{sec:stable_grasp_study_details} provides the stable grasp study on ARCTIC and HOI4D dataset.
\cref{sec:eval_ihoi} showcases failure of current CAD-agnostic approach on EPIC-Grasps dataset.
\cref{sec:ablation_hand_pose_method} provides ablation of the hand pose estimation method.
\cref{sec:further_impl} provides further implementation details.

\section{Statistics of EPIC-Grasps annotation}
\label{sec:epic_grasp_stats}

In Table~\ref{tab:ds_stats}, we present average, standard deviation and maximum clip duration for each of our categories as well as the total number of video clips per category in EPIC-Grasps.
Note that grasping some categories (e.g. pan) can be significantly longer in duration than others (e.g. mug). 

\section{Statistics of ARCTIC-Grasps automatic annotation}
\label{sec:arctic_grasp_stats}

We use Eq~(2) in the main paper to automatically extract stable grasp duration from the 3D ARCTIC~\cite{fan2023arctic} dataset, 
where in-contact area can be computed from ground truth meshes.
We set $\tau$ to 0.5 empirically: we inspect the resulting grasps to ensure the start-end time match visual expectation. 
As our pipeline assumes rigid object, we exclude segments where the hand is changing the object's articulation.

In Table~\ref{tab:arctic_grasps_stats}, we present average, standard deviation and maximum clip duration for each category as well as total number of video clips per category in ARCTIC-Grasps.
As ARCTIC sequences do not include actual activities (e.g. a box is opened only to be closed again rather than to retrieve something from inside), the stable grasps are shorter than those in the unscripted functional recordings of EPIC-Grasps.

\begin{table}[t]
    \centering
    \begin{tabular}{c | c c c | c }
    & \multicolumn{3}{c|}{Duration (sec)} & \\
     Category  & avg & std. & max & \#-clips\\
         \hline
    bottle & 4.25 & 4.70 & 29.40 & 311\\
    bowl & 3.59 & 5.53 & 43.36 & 421\\
    can & 3.50 & 4.11 & 22.76 & 105\\
    cup & 2.79 & 3.20 & 19.72 & 125\\
    glass & 2.45 & 3.73 & 34.72 & 226\\
    mug & 2.40 & 3.34 & 29.74 & 167\\
    pan & 6.39 & 9.36 & 91.28 & 526\\
    plate & 2.77 & 4.49 & 54.60 & 473\\
    saucepan & 5.73 & 10.08 & 76.10 & 77\\
    \hline
    All & 3.95 & 6.30 & 91.28 & 2431\\
    \end{tabular}
    \caption{EPIC-Grasps sequences statistics}
    \label{tab:ds_stats}
\end{table}

\begin{table}[t]
    \centering
    \begin{tabular}{c | c c c | c }
    & \multicolumn{3}{c|}{Duration (sec)} & \\
    Category  & avg & std. & max & \#-clips\\
    \hline
    box & 1.52 & 0.75 & 4.23 & 138\\
    capsulemachine & 1.51 & 0.69 & 3.60 & 95\\
    espressomachine & 1.72 & 0.95 & 4.73 & 101\\
    ketchup & 1.51 & 0.80 & 4.83 & 106\\
    laptop & 1.39 & 0.74 & 4.90 & 144\\
    microwave & 1.51 & 0.81 & 5.53 & 112\\
    mixer & 1.44 & 0.68 & 3.30 & 122\\
    notebook & 1.15 & 0.44 & 2.83 & 151\\
    phone & 1.30 & 0.91 & 8.07 & 146\\
    scissors & 1.60 & 0.77 & 4.37 & 57\\
    waffleiron & 1.21 & 0.58 & 3.90 & 131\\
    \hline
    All & 1.42 & 0.76 & 8.07 & 1303\\
    \end{tabular}
    \caption{ARCTIC-Grasps sequences statistics}
    \label{tab:arctic_grasps_stats}
\end{table}

\section{Stable Grasp Study on ARCTIC and HOI4D datasets.}
\label{sec:stable_grasp_study_details}

\begin{figure}[t]
    \centering
    \includegraphics[width=\textwidth]{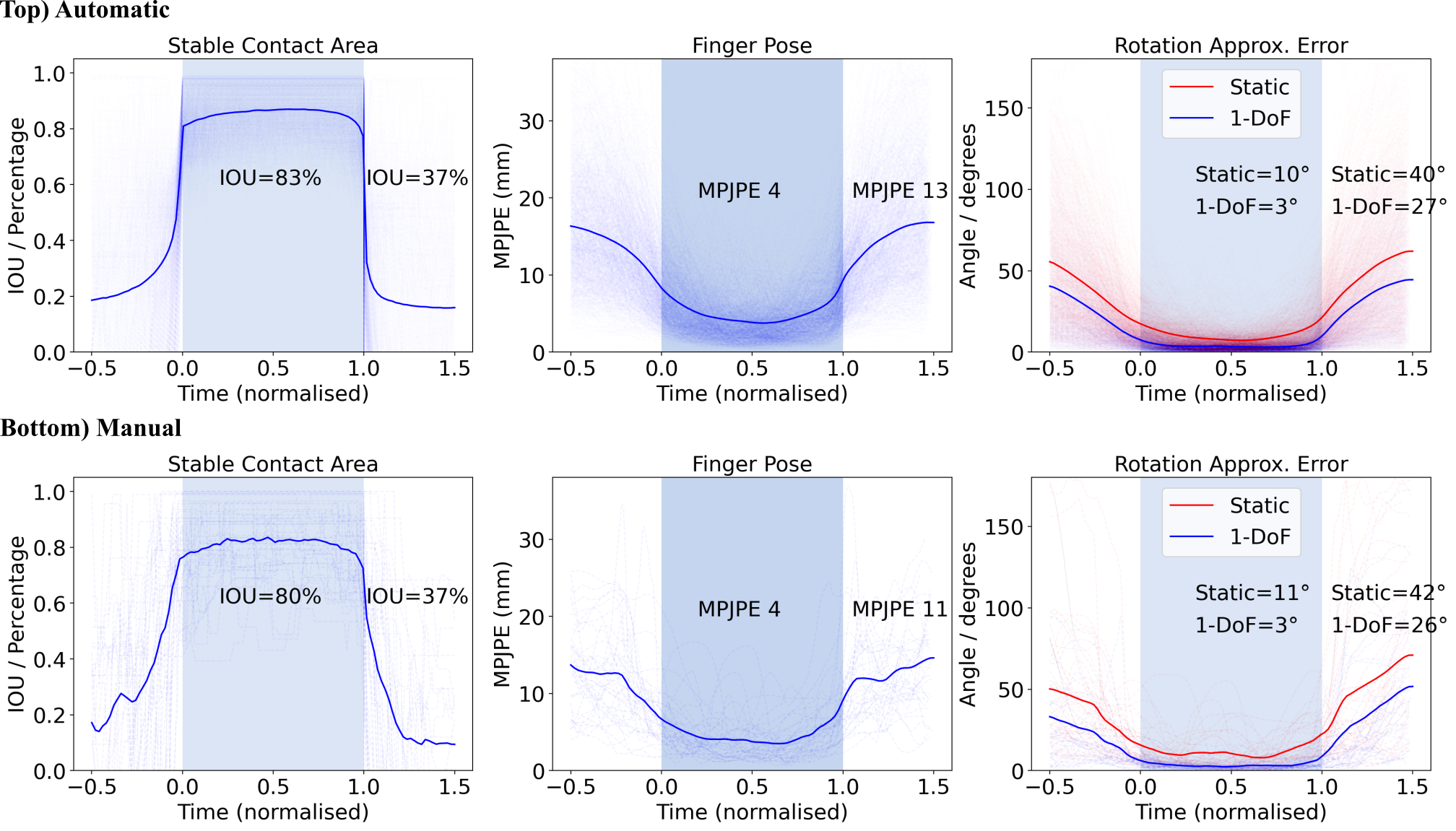}
    \caption{
    Metrics on ARCTIC within/outside stable grasps: Comparing object in-contact area (left), finger pose (middle)
    and relative object rotation w.r.t. the hand.
    Normalising all stable grasp duration for direct comparison.
    }
    \label{fig:analysis_arctic}
\end{figure}

\begin{figure}[t]
    \centering
    \includegraphics[width=\textwidth]{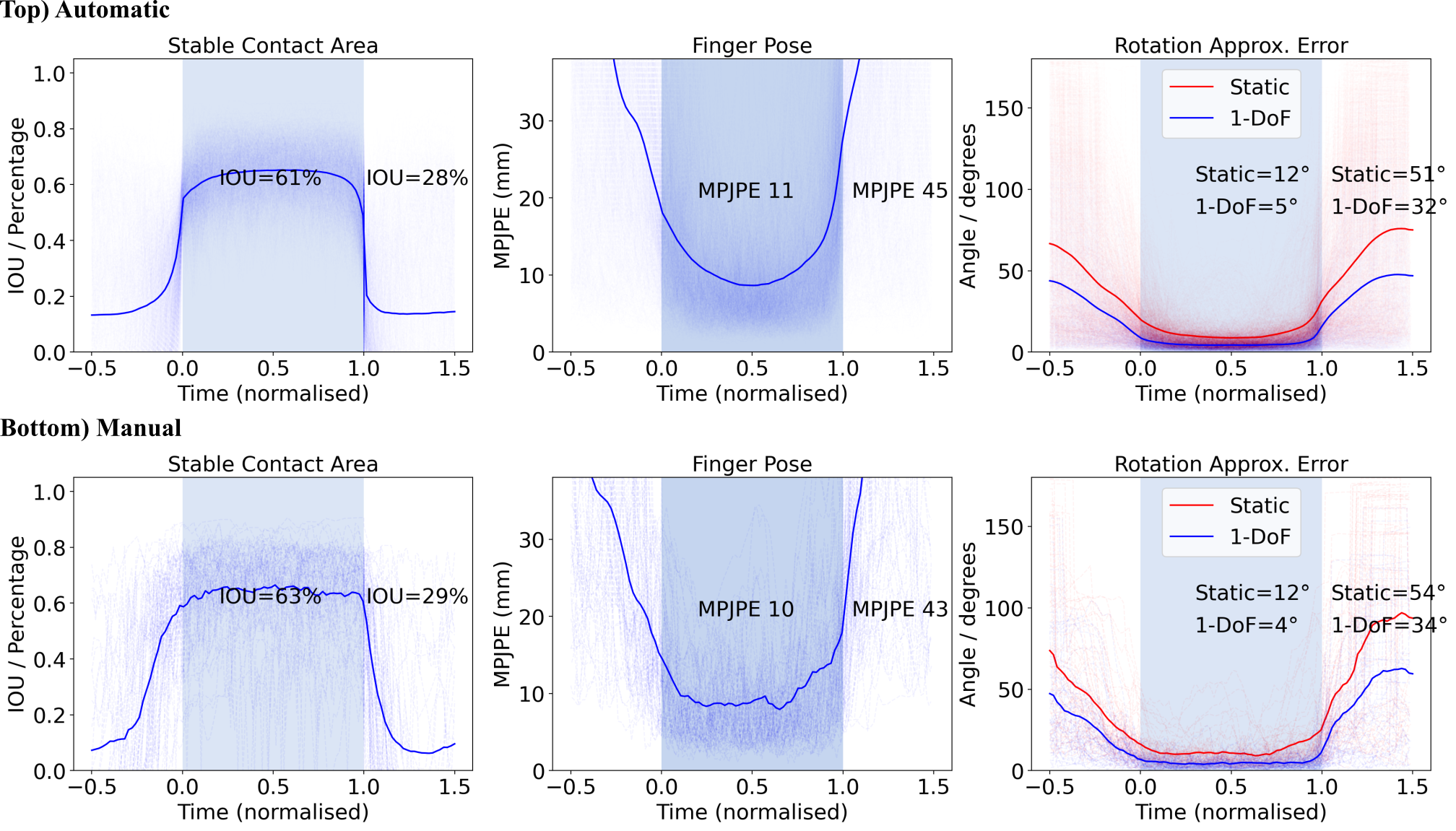}
    \caption{
    Metrics on HOI4D within/outside stable grasps: Comparing object in-contact area (left), finger pose (middle)
    and relative object rotation w.r.t. the hand.
    Normalising all stable grasp duration for direct comparison.
    }
    \label{fig:analysis_hoi4d}
\end{figure}

In this section, we analyse metrics related to stable grasps:
stable contact area, finger pose variations and object rotation within/outside stable grasps.
Note that hand wrist pose changes drastically w.r.t. the camera, as shown in Fig~3 in the main paper. 

\noindent \textbf{ARCTIC Dataset.} 
In~\cref{fig:analysis_arctic} (top),
we plot stable contact area, finger pose variations and object rotation errors on 1303 automatically extracted sequences from ARCTIC-Grasps.
Among these three metrics, the stable contact area (top-left) shows the sharpest change within/outside stable grasps.
We demonstrate that both the finger poses and the object relative orientation do change during the stable grasps, although their changes are bigger outside grasps.
Note that the top-left and top-right plots in  \cref{fig:analysis_arctic} are the same as those in Fig~4 of the main paper. 

Furthermore, we manually annotate 40 stable grasps in ARCTIC dataset, by observing the videos solely without using the 3D ground truth. We plot the three metrics in~\cref{fig:analysis_arctic} (bottom). 
As expected, \cref{fig:analysis_arctic} (bottom-left) shows that the stable grasp duration corresponds to a consistent in-contact area for manual annotated stable grasps.
We also verify that the rotation approximation error of 1-DoF are smaller than Static for manually obtained annotations.
These figures confirm that the automatically extracted stable gasps are a good approximation of manual annotations of stable grasps.

\noindent \textbf{HOI4D Dataset.}
To showcase our assumptions on stable grasp generalise to other datasets with 3D annotations, we replicate the analysis above, % on ARCTIC 
on the publicly available HOI4D dataset~\cite{Liu2022HOI4D:Interaction}.
We use the Eq~(2) again to obtain 3568 stable grasps sequences from the HOI4D.
Note that the noise of the 3D ground truth pose in HOI4D are much higher than ARCTIC, we thus increase the contact distance threshold to 1cm (from 0cm in ARCTIC) and
set $\tau$ to a lower value 0.3.
As in ARCTIC, we also manually annotate 60 stable grasps in HOI4D dataset, by observing the videos, and compute the same metrics.

In~\cref{fig:analysis_hoi4d}, 
we compare the three metrics on the stable grasp sequences on HOI4D,
for both automatic annotations (top) and manually annotated samples (bottom).
As expected, 
the stable contact area shows sharp drops at the temporal boundary of stable grasps. 
Similarly, we observed that the object do rotate relative to the hand, while the rotation error off the 1-DoF axis are much smaller.

These figures showcase that:
\begin{itemize}
    \item Our definition of the stable contact temporal duration extends to multiple 3D-annotated datasets.
    \item On a variety of datasets, subjects and manipulated objects, the finger poses exhibit considerable changes during the stable grasp.
    \item On both datasets and all sequences, objects rotate within the stable grasp -- the static assumption produces a large error within the stable grasp ($11^\circ$ in ARCTIC compared to $12^\circ$ in HOI4D).
    \item Optimising for the rotation around a latent axis, the error off the 1-DoF assumption is significantly smaller and better corresponds to the stable grasp temporal extent. This conclusion is verified on both datasets and both manual and automatically annotated sequences.
\end{itemize}

\section{Evaluating CAD-agnostic method IHOI}
\label{sec:eval_ihoi}

In our method, we assume knowledge of the CAD model.
We also explored the recent works that attempt hand grasps without such knowledge.
In this section, we showcase these models to be unusable for in-the-wild stable grasp reconstruction through three examples, qualitatively.

\begin{figure}[t]
    \centering
    \includegraphics[width=0.5\linewidth]{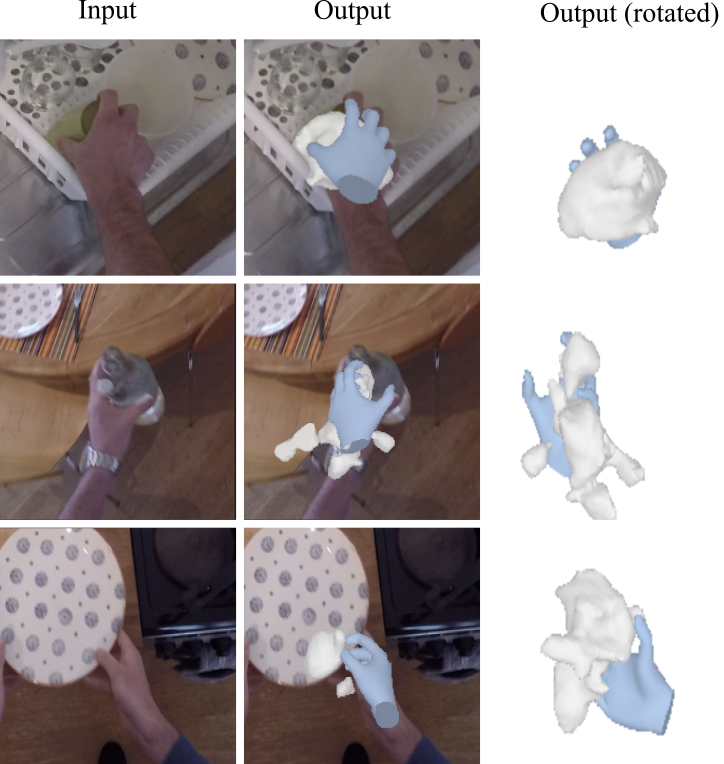}
    \caption{In-the-wild test of IHOI~\cite{YeWhatsHands}}
    \label{fig:Ihoi}
\end{figure}

The shape reconstruction method proposed in~\cite{Huang2022ReconstructingVideo} requires careful scan of the underlying in-hand object hence is not suitable for testing in our case. 
We also note that HOLDNet~\cite{fan2024hold} and DiffHOI~\cite{ye2023vhoi} do not have released code to be used.
Therefore, we evaluate the image-based CAD-agnostic method IHOI~\cite{YeWhatsHands} on the EPIC-Grasps dataset, using a model trained on HO3D~\cite{HampaliHOnnotate:Poses} images that contain the same categories: mug and bottle -- i.e. the method is aware of these shapes during training. 

\cref{fig:Ihoi} shows IHOI is unable to generate plausible object meshes.
As fingers are mostly occluded in egocentric views, this makes the challenge of reconstruction significantly higher than side-view reconstructions where fingers are mostly visible.
While CAD-agnostic reconstruction is the ultimate goal, these approaches are currently far from being useful for our scenario.

\section{Ablation on hand pose estimation method}
\label{sec:ablation_hand_pose_method}

While we use the recent transformer-based approach for estimating hand-pose, HaMeR~\cite{pavlakos2024reconstructing}, we test our method using the FrankMocap baseline for hand pose etimation~\cite{RongFrankMocap:Integration}. This is an important ablation as our baseline uses FrankMocap.

In~\cref{tab:ablate_hand_pose_estimation}, we showcase that our method, using FrankMocap~\cite{RongFrankMocap:Integration} for hand pose etimation outperforms the baseline HOMan~\cite{Hasson2021TowardsVideos} which also uses FrankMocap. Using the same hand initialisation, we outperform HOMan~\cite{Hasson2021TowardsVideos} by a wide margin - SCA@0.6 improves from 3.3 to 20.3.
Using HaMeR, the method further improves consistently over all categories. 
Average SCA@0.6 improves by an additional 12.6\% to reach 32.9.

\begin{figure}[t]
    \centering
    \includegraphics[width=0.8\textwidth]{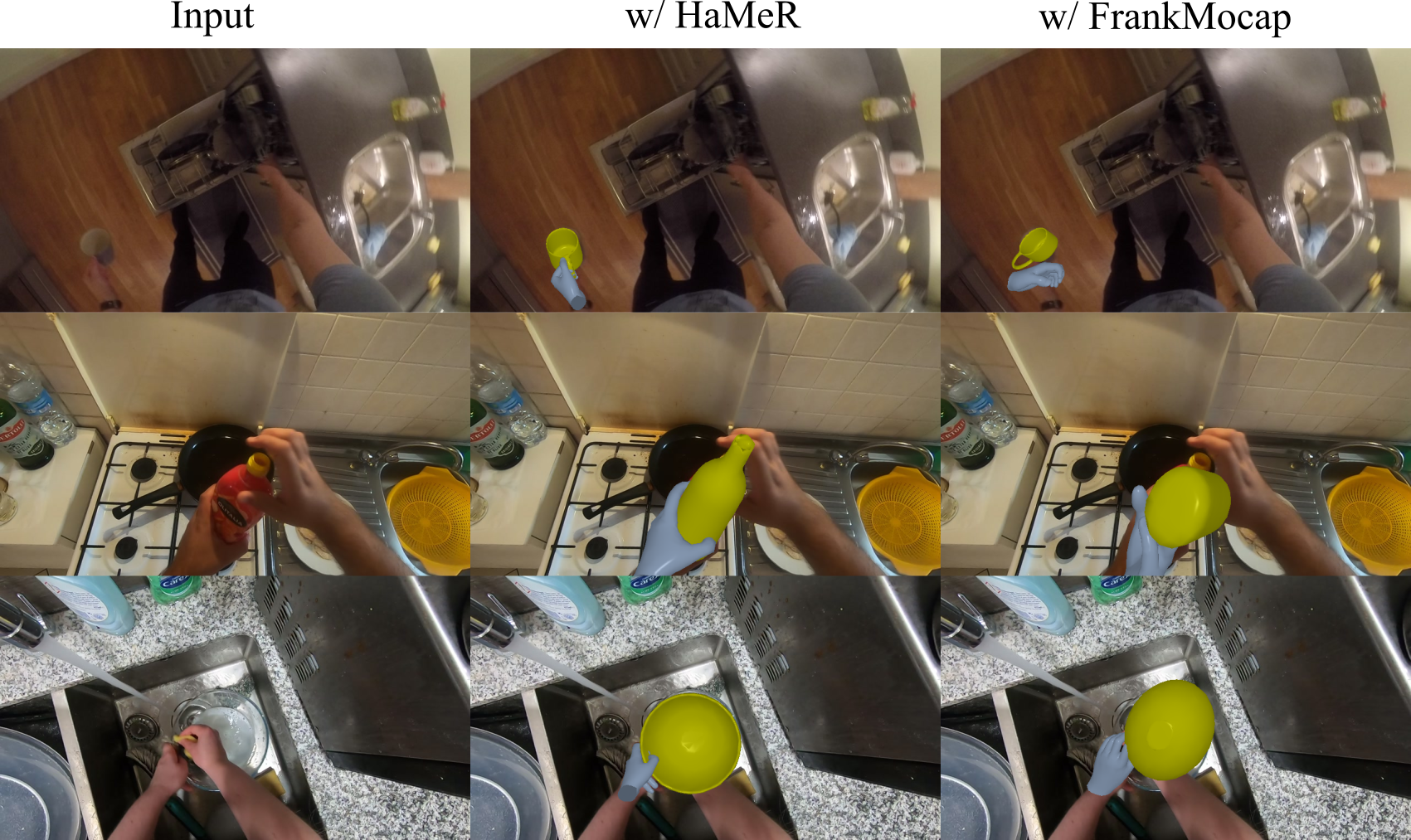}
    \caption{Qualitative comparison between HaMeR and FrankMocap, when used with our method.}
    \label{fig:ablate_hand}
\end{figure}

\cref{fig:ablate_hand} shows that FrankMocap struggles with the hand orientation, 
resulting incorrect reconstruction results.

\section{Further Implementation Details}
\label{sec:further_impl}

\noindent \textbf{Physical Heuristics}
%%%%%% Physical losses
In Eq~(5) of the main paper we note our usage of physical attraction and repulsion losses $E_{push}$ and $E_{pull}$.

\begin{figure}[t]
    \centering
    \includegraphics[width=0.2\linewidth]{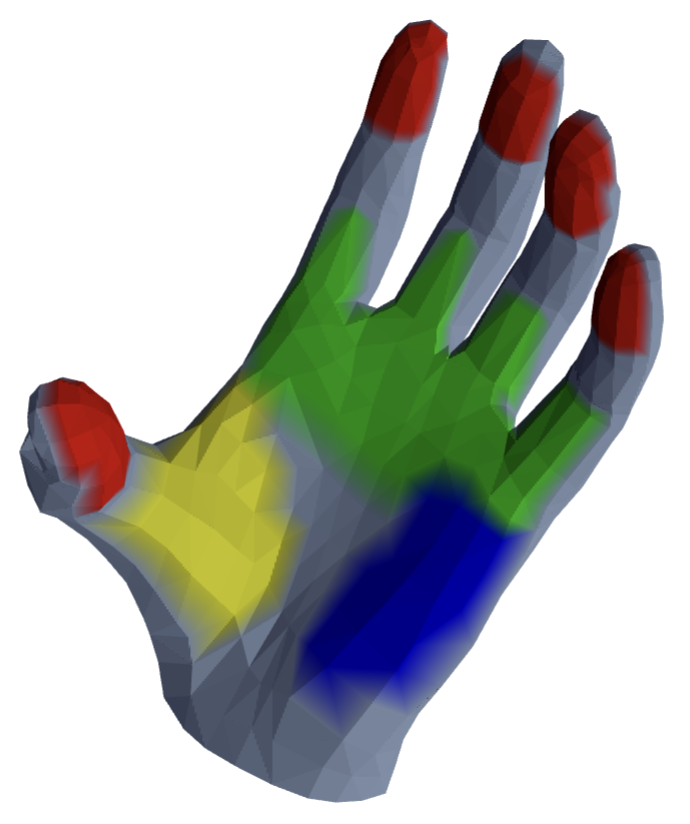}
    \caption{Eight contact regions: five finger tips + three palm areas. The contact regions serve two purposes: bounding the object inside and attracting the object closer to these regions.}
    \label{fig:handregions}
\end{figure}

The term $E_{push}$ ensures  all object vertices are located inside 
the contact surface of the hand (Fig~\ref{fig:handregions}).
For each $v_o \in V_o$, 
we locate the nearest vertex in hand contact regions,
and compute distance along the surface normal of this hand vertex.
Object vertices that penetrate into the contact surface will have negative values.
We maximise those negative values, truncating the positive ones:
\begin{align}
\label{eq:inside}
\Indexed{E_{push}} &= \sum_{v_o \in V_o} -1 * \min( d_v, 0 )\\
d_v &= \langle v_o - v^*_h, n^*_h \rangle 
\label{eq:inside_dv}
\end{align}
where $v^{*}_h$ is the corresponding nearest vertex on the hand and $n^{*}_h$ is the surface normal of $v^{*}_h$.

In addition to $E_{push}$, which pushes the object out of the penetrating region against the hand, we use a balancing loss $E_{pull}$ which pulls the object to touch these contact regions.
We here focus on the eight contact regions showcased in Fig~\ref{fig:handregions}.
For each contact region with hand vertices $\{v_h\}_C$, the region-to-object distance is defined as the minimum distance of all $(v_h, v_o)$ pairs. We use 5 finger tip regions and minimise the average of these region-to-object distances.
\begin{align}
\Indexed{E_{pull}} & = \frac{1}{5}\sum_{C}d(\{v_h\}_C, V_o) \\
d(\{v_h\}_C, V_o) & = \min_{v_h \in \{v_h\}_C, v_o \in V_o} \langle v_h - v_o, n_o \rangle
\label{eq:close}
\end{align}
where $n_o$ is the surface normal of $v_o$.

\begin{table}[t]
    \centering
    \resizebox{0.8\textwidth}{!}{%
\begin{tabular}{l|ccc|ccc|ccc}
    \toprule
     & \multicolumn{3}{c|}{HOMan~\cite{Hasson2021TowardsVideos}} & \multicolumn{3}{c|}{1-DoF + FrankMocap} & \multicolumn{3}{c}{1-DoF + HaMeR} \\
     Category & IOU & SCA@0.8 & SCA@0.6 & IOU & SCA@0.8 & SCA@0.6 & IOU & SCA@0.8 & SCA@0.6 \\
    \midrule
    bottle & 56.7 & 3.4 & 5.8 & \cellcolor{yellow!25}66.7 & \cellcolor{yellow!25}11.7 & \cellcolor{yellow!25}36.8 & \cellcolor{green!25}72.1 & \cellcolor{green!25}26.1 & \cellcolor{green!25}51.3 \\
    bowl & \cellcolor{yellow!25}54.4 & 1.1 & 1.6 & 53.5 & \cellcolor{yellow!25}5.5 & \cellcolor{yellow!25}14.4 & \cellcolor{green!25}56.2 & \cellcolor{green!25}11.1 & \cellcolor{green!25}25.7 \\
    can & 48.3 & 3.4 & 5.1 & \cellcolor{yellow!25}53.1 & \cellcolor{yellow!25}11.0 & \cellcolor{yellow!25}20.9 & \cellcolor{green!25}54.1 & \cellcolor{green!25}16.5 & \cellcolor{green!25}26.6 \\
    cup & 56.9 & 5.5 & 7.8 & \cellcolor{yellow!25}64.2 & \cellcolor{yellow!25}11.5 & \cellcolor{yellow!25}32.5 & \cellcolor{green!25}65.9 & \cellcolor{green!25}18.4 & \cellcolor{green!25}46.1 \\
    glass & 55.4 & 3.0 & 4.2 & \cellcolor{yellow!25}58.8 & \cellcolor{yellow!25}10.7 & \cellcolor{yellow!25}25.7 & \cellcolor{green!25}62.5 & \cellcolor{green!25}15.7 & \cellcolor{green!25}37.8 \\
    mug & 59.4 & 3.9 & 6.6 & \cellcolor{yellow!25}61.1 & \cellcolor{yellow!25}7.0 & \cellcolor{yellow!25}28.3 & \cellcolor{green!25}62.6 & \cellcolor{green!25}10.8 & \cellcolor{green!25}38.6 \\
    pan & \cellcolor{yellow!25}48.3 & 0.5 & 1.9 & 39.7 & \cellcolor{yellow!25}2.0 & \cellcolor{yellow!25}6.9 & \cellcolor{green!25}49.1 & \cellcolor{green!25}7.0 & \cellcolor{green!25}20.4 \\
    plate & \cellcolor{yellow!25}61.1 & 1.0 & 1.6 & 61.0 & \cellcolor{yellow!25}9.0 & \cellcolor{yellow!25}21.5 & \cellcolor{green!25}65.2 & \cellcolor{green!25}16.3 & \cellcolor{green!25}34.6 \\
    saucepan & 51.1 & 0.4 & 2.4 & \cellcolor{yellow!25}51.4 & \cellcolor{yellow!25}1.8 & \cellcolor{yellow!25}16.2 & \cellcolor{green!25}56.8 & \cellcolor{green!25}8.5 & \cellcolor{green!25}34.6 \\
    \hline
    All & 54.9 & 1.9 & 3.3 & \cellcolor{yellow!25}55.1 & \cellcolor{yellow!25}7.2 & \cellcolor{yellow!25}20.3 & \cellcolor{green!25}\textbf{59.9} & \cellcolor{green!25}\textbf{14.1} & \cellcolor{green!25}\textbf{32.9} \\
    \bottomrule
    \end{tabular}    
}
    \caption{Ablation on the hand pose estimation method. 
    \colorbox{green!25}{Green} shows the \colorbox{green!25}{best} per metric,
    \colorbox{yellow!25}{Yellow} shows the \colorbox{yellow!25}{second} best.
    }
    \label{tab:ablate_hand_pose_estimation}
\end{table}

\noindent \textbf{Initialisation of poses on EPIC-Grasps.}
For EPIC-Grasps dataset, we manually set initial object relative poses to the common poses of each category, results in 80 total initial poses for 9 object categories.

\noindent \textbf{Initialisation of poses on ARCTIC-Grasps.}
For ARCTIC-Stable dataset, the initial rotations are generated by clustering the ground-truth rotations, where clustering is performed via the axis-angle representation of the rotation matrix. We initialise 50 rotations and 1 global translation for each (object, left/right hand) pair.

\end{document}